\documentclass[10pt]{article}
\usepackage[a4paper, total={6in, 8in}]{geometry}
\usepackage[utf8]{inputenc}
\usepackage[T1]{fontenc}
\usepackage{authblk}
\usepackage{amsmath}
\usepackage{url}

\usepackage{rotating}

\date{}

\title{Explaining the Unique Nature of Individual Gait Patterns with Deep Learning\footnotetext{This is a pre-print of an article
published in \emph{Scientific Reports}. The final authenticated version
is available online at: \url{https://doi.org/10.1038/s41598-019-38748-8}}}

\author[1,+]{Fabian Horst}
\author[2,+]{Sebastian Lapuschkin}
\author[2,*]{Wojciech Samek}
\author[3,4,5,*]{\mbox{Klaus-Robert Müller}}
\author[1,*]{Wolfgang I. Schöllhorn}

\affil[1]{Department of Training and Movement Science, Institute of Sport Science, Johannes Gutenberg-University Mainz, Mainz, Rhineland-Palatinate, Germany}
\affil[2]{Department of Video Coding \& Analytics, Fraunhofer Heinrich Hertz Institute, Berlin, Germany}
\affil[3]{Department of Electrical Engineering \& Computer Science, Technical University Berlin, Berlin, Germany}
\affil[4]{Department of Brain and Cognitive Engineering, Korea University, Seoul, Korea}
\affil[5]{Max Planck Institute for Informatics, Saarbrücken, Saarland, Germany}
\affil[*]{wojciech.samek@hhi.fraunhofer.de (WS)}
\affil[*]{klaus-robert.mueller@tu-berlin.de (KRM)}
\affil[*]{wolfgang.schoellhorn@uni-mainz.de (WIS)}
\affil[+]{these authors contributed equally to this work}

\begin{document}
\flushbottom
\maketitle
\begin{abstract}
    Machine learning (ML) techniques such as (deep) artificial neural networks (DNN) are solving very successfully a plethora of tasks and provide new predictive models for complex physical, chemical, biological and social systems. However, in most cases this comes with the disadvantage of acting as a black box, rarely providing information about what made them arrive at a particular prediction. This black box aspect of ML techniques can be problematic especially in medical diagnoses, so far hampering a clinical acceptance. The present paper studies the uniqueness of individual gait patterns in clinical biomechanics using DNNs. By attributing portions of the model predictions back to the input variables (ground reaction forces and full-body joint angles), the Layer-Wise Relevance Propagation (LRP) technique reliably demonstrates which variables at what time windows of the gait cycle are most relevant for the characterisation of gait patterns from a certain individual. By measuring the time-resolved contribution of each input variable to the prediction of ML techniques such as DNNs, our method describes the first general framework that enables to understand and interpret non-linear ML methods in (biomechanical) gait analysis and thereby supplies a powerful tool for analysis, diagnosis and treatment of human gait.
\end{abstract}

\newcommand{\eg}{e.g.\ }
\newcommand{\ie}{i.e.\ }
\newcommand{\x}{\mathbf{x}}
\newcommand{\w}{\mathbf{w}}
\renewcommand{\b}{\mathbf{b}}
\newcommand{\todo}[1]{{\color{teal} #1}}

\section*{Introduction}
The ability to walk is crucial for human mobility and enables to predict quality of life, morbidity and mortality
\cite{verghese2006epidemiology, verghese2012mobility, soh2011determinants, studenski2003physical, studenski2011gait, hirsch2011predicting, fagerstrom2010mobility, mahlknecht2013prevalence, rubenstein2001quality, forte2015health}.
Its importance is underlined by the fear of losing the ability to walk, which is frequently considered to be the first and most significant concern from individuals that sustain diagnoses like stroke
\cite{schmid2007improvements, seale2010gait} or Parkinson disease \cite{soh2011determinants, ellis2011measures}.
However, gait and balance are no longer regarded as purely motor tasks, but are considered as complex sensorimotor behaviours that are heavily affected by cognitive and affective aspects \cite{giladi2013classification}.
This may partially explain the sensitivity to subtle neuronal dysfunction, and why gait and postural control can predict the development of disease such as diabetes, dementia or Parkinson even years before they are diagnosed clinically \cite{giladi2013classification, verghese2002abnormality, baltadjieva2006marked, buracchio2010trajectory, valkanova2017can}.

In order to prevent, diagnose, or rehabilitate a loss of independence due to (gait) impairments, gait analysis is common practice to support and standardise researchers’, clinicians’ and therapists’ decisions when assessing gait abnormalities and/or identifying changes due to orthopaedic or physiotherapeutic interventions \cite{baker2013measuring}.
But although becoming gradually established over the past decades, most biomechanical gait analyses have examined the influence of single time-discrete gait variables, like gait velocity, step length or range of motion, as risk factor or predictor for (gait) disease in isolation\cite{mills2013biomechanical, wegener2011effect}.
While conventional approaches have addressed successfully many important clinical and scientific questions related to human gait (impairments), they exhibit some inherent limitations:
When single time-discrete variables (\eg the range of motion in the knee joint) are extracted from time-continuous variables (\eg knee joint angle-gait stride curve), a large amount of data are discarded.
In many cases it remains unclear, if and to what degree single pre-selected variables are capable to represent a sufficient description of a whole body movement like human gait \cite{schollhorn2002identification, federolf2012holistic, eskofier2013marker, chau2001review1, chau2001review2}.
The a priori selection of single gait variables relies mostly on the experience and/or subjective opinion of the analyst, which may lead to a certain risk of investigator bias.
Furthermore, single pre-selected variables might miss potentially meaningful information that are represented by -- or in combination with -- other (not selected) variables.
In this context, it seems questionable if the multi-dimensional interactions between gait characteristics and gait disease or disease that impair gait can be entirely represented by a subjective selection of single time-discrete variables \cite{schollhorn2002identification, federolf2012holistic, schollhorn2004applications}.

In response to these shortcomings, multivariate statistical analysis \cite{chau2001review1, chau2001review2, wolf2006automated} and machine learning techniques such as artificial neural networks (ANN)\cite{bishop1995neural} and support vector machines (SVM)\cite{boser1992training, cortes1995support, muller2001introduction, scholkopf2002learning} have been used to examine human locomotion based on time-continuous gait patterns \cite{schollhorn2004applications}.
Significant advances in motion capture equipment and data analysis techniques have enabled a plethora of studies that have advanced our understanding about human gait.
Due to extensive new datasets \cite{mckay20161000}, the application of machine learning techniques is becoming increasingly popular in the area of clinical biomechanics \cite{schollhorn2004applications, phinyomark2018analysis, figueiredo2018automatic} and provided new insights into the nature of human gait control.
The application of ANNs and SVMs highlighted for example that gait patterns are unique to an individual person \cite{schollhorn2002identification, horst2017one}, exhibited natural changes within different time-scales \cite{horst2016daily, horst2017intra} and identified that emotional states \cite{janssen2008recognition} and grades of fatigue \cite{janssen2011diagnosing} can be differentiated from human gait patterns.
Furthermore, gender and age-specific gait patterns could be differentiated \cite{eskofier2013marker, begg2005machine}.
In the context of clinical gait analysis, several approaches based on machine learning have been published in recent years in order to support clinicians in identifying and categorizing specific gait patterns into clinically relevant groups \cite{chau2001review1, chau2001review2, schollhorn2004applications, figueiredo2018automatic}.
Previous studies were able to differentiate gait patterns from healthy individuals and individuals with (neurological) disorders like Parkinson’s disease \cite{zeng2016parkinson}, cerebral palsy \cite{alaqtash2011automatic}, multiples scleroses \cite{alaqtash2011automatic} or traumatic brain injuries \cite{williams2015classification} and pathological gait conditions like lower-limb fractures \cite{figueiredo2018automatic} or acute anterior cruciate ligament injury \cite{christian2016computer}.

Although machine learning techniques are solving very successfully a variety of classification tasks and provide new insights from complex physical, chemical, biological, or social systems, in most cases they go along with the disadvantage of acting as a black box, rarely providing information about what made them arrive at a particular decision \cite{baehrens2010explain,montavon2018methods}.
This non-transparent operating and decision-making of most non-linear machine learning methods leads to the problem that their predictions are not straightforward understandable and interpretable.
 This black box manner can be problematic especially in applications of machine learning in medical diagnosis like gait analyses and so far strongly hamper their clinical acceptance \cite{wolf2006automated}.
The lack of understanding and interpreting the decision process of machine learning techniques is a clear drawback and recently attracted attention in the field of machine learning
\cite{baehrens2010explain,montavon2018methods,gevrey2003review,simonyan2013deep,zeiler2014visualizing, BachPLOS15, zintgraf2016new, ribeiro2016should, zhang2016top, selvaraju2017grad, fong2017interpretable, MonPR17}.
In this context, the so called Layer-wise Relevance Propagation (LRP) technique has been proposed as general technique for explaining classifier's decisions by decomposition, i.e.\ by measuring the contribution of each input variable to the overall prediction  \cite{BachPLOS15}.
LRP has been successfully applied to a number of technical and scientific tasks such as image classification \cite{SamTNNLS17, LapAMFG17} and text document classification \cite{arras2017relevant}.
Also, interpreting linear and non-linear models have helped to gain interesting insights in neuroscience \cite{haufe2014interpretation, StuJNM16}, bioinformatics \cite{zien2009feature,stegle2012using, vidovic2015svm2motif} and physics \cite{schutt2017quantum}.

Due to benefits of machine learning methods in comparison to conventional approaches in gait analysis \cite{schollhorn2004applications, phinyomark2018analysis}, LRP appears highly promising to increase their transparency and therefore make them applicable and reliable for clinical diagnoses \cite{chau2001review1, chau2001review2, wolf2006automated}.
In the context of personalised medicine, the aim of the present study was to examine individual gait patterns by:
\begin{enumerate}
\item demonstrating the uniqueness of gait patterns to the individual by using (deep) artificial neural networks for predicting identities based on gait;
\item verifying that non-linear machine learning methods such as (deep) artificial neural networks use comprehensible prediction strategies and learn meaningful gait characteristics by using the Layer-Wise Relevance Propagation; and
\item analysing the unique gait signature from an individual by highlighting which variables at what time windows of the gait cycle are used by the model to identify an individual.
\end{enumerate}

The presented approach investigates the suitability of understanding and interpreting the classification of gait patterns using state-of-the-art machine learning methods.
This paper therefore presents a first step towards establishing a powerful tool that can be used as the basis for future application of machine learning in (biomechanical) gait analysis and thus enabling automatic classifications of (neurological) disorders and pathological gait conditions applicable in clinical diagnoses.

\section*{Results}

\begin{sidewaysfigure}
    \begin{center}
    \includegraphics[width=1.\textwidth]{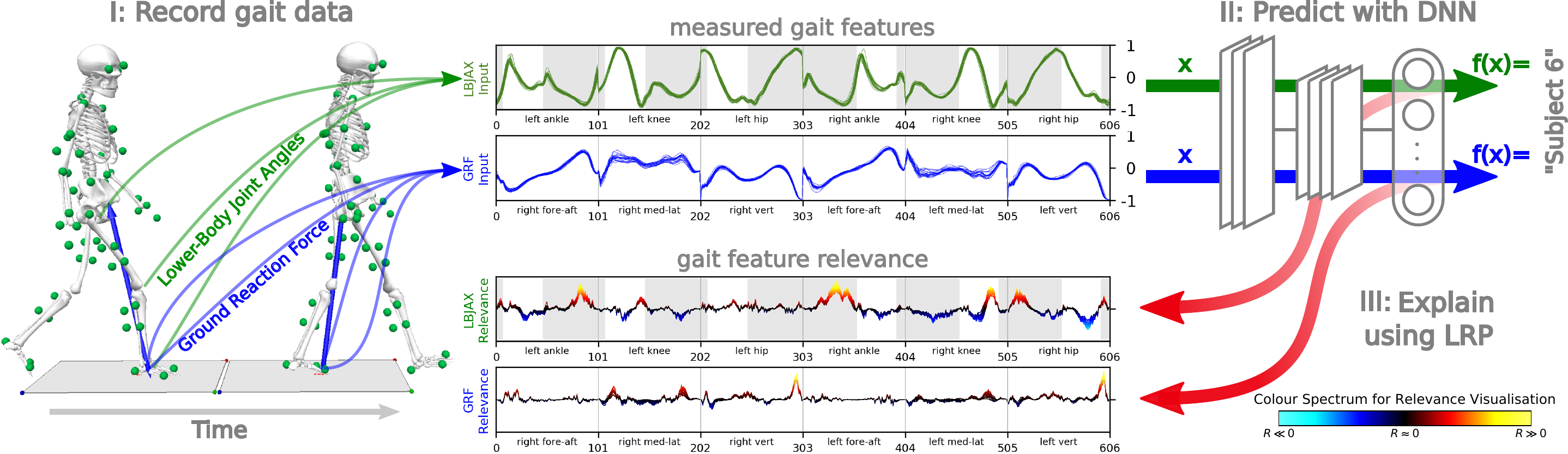}
    \caption{Overview of data acquisition and data analysis, showing the example of subject 6.
    \emph{I:} The biomechanical gait analysis compromised the recording of 20 times walking barefoot a distance of 10 m at a self-selected walking speed.
    Two force plates and ten infrared cameras recorded the three-dimensional full-body joint angles and ground reaction forces during a double step.
    \emph{II:} Lower-body joint angles in the sagittal plane (flexion-extension) (LBJAX) and ground reaction forces (GRF) compromising the fore-aft shear force (fore-aft), medial-lateral shear force (med-lat) and vertical force (vert) have been used as time-normalised and concatenated input vectors $\x$ for the prediction of subjects $y$ using deep artificial neural networks (DNN). Shaded areas for the LBJAX highlight the time where the respective (left or right) foot is in contact with the ground.
    \emph{III:} Decomposition of input relevance values using the Layer-Wise Relevance Propagation (LRP).
    Colour Spectrum for the visualisation of input relevance values of the model predictions.
    Throughout this manuscript, we use LRP to exclusively analyse the prediction for the true class of a sample.
    Thereby, black line segments are irrelevant to the model’s prediction. Red and hot colours identify input segments causing a prediction corresponding to the subject label, while blue and cold hues are features contradicting the subject label.
    For subject 6, the predicting model (CNN-A) achieves true positive rates (TP) of $100\%$ for LBJAX and $95.23\%$ for GRF.}
    \label{fig:overview}
    \end{center}
\end{sidewaysfigure}

The uniqueness of human walking to the individual was examined based on time-continuous kinematic (full-body joint angles) and kinetic (ground reaction forces) gait patterns (see Methods section for a description of the data).
From a biomechanical gait analysis (Figure~\ref{fig:overview} \emph{I: Record gait data}), conducted on 57 healthy subjects, lower-body joint angles (LBJA) and ground reaction forces (GRF) have been measured as input vectors $\x$ for the prediction of subjects $y$ using deep artificial neural networks (DNN) (Figure \ref{fig:overview} \emph{II: Predict with DNN}).
LRP decomposes the prediction $f(\x)$ of a learned function $f$ given an input sample $\x$ into into time-resolved input relevance values $R_i$ for each time-discrete input $x_i$, which enables to explain the prediction of DNNs as partial contributions from individual input components (Figure \ref{fig:overview} \emph{III: Explain prediction using LRP}).
LRP indicates based on which information a model predicts and thereby enables to interpret the input relevance values and their dynamics as representation for a certain class (individual).
Hence, the input relevance values point out which gait characteristics were most relevant for the identification of a certain individual.
In the following, input relevance values are visualised by colour coding, using a diverging and symmetric high contrast colour scheme as shown in Figure \ref{fig:overview} (\emph{III: Explain prediction using LRP}).
Here, input elements neutral to the predictor ($R_i \approx 0$) will be shown in black colour, while warm and hot hues indicate input components supporting the prediction ( $R_i \gg 0 $ ) of the analysed class and cold hues identify contradictory inputs ( $R_i \ll 0 $ ).

\begin{figure}
    \begin{center}
    \includegraphics[width=\textwidth]{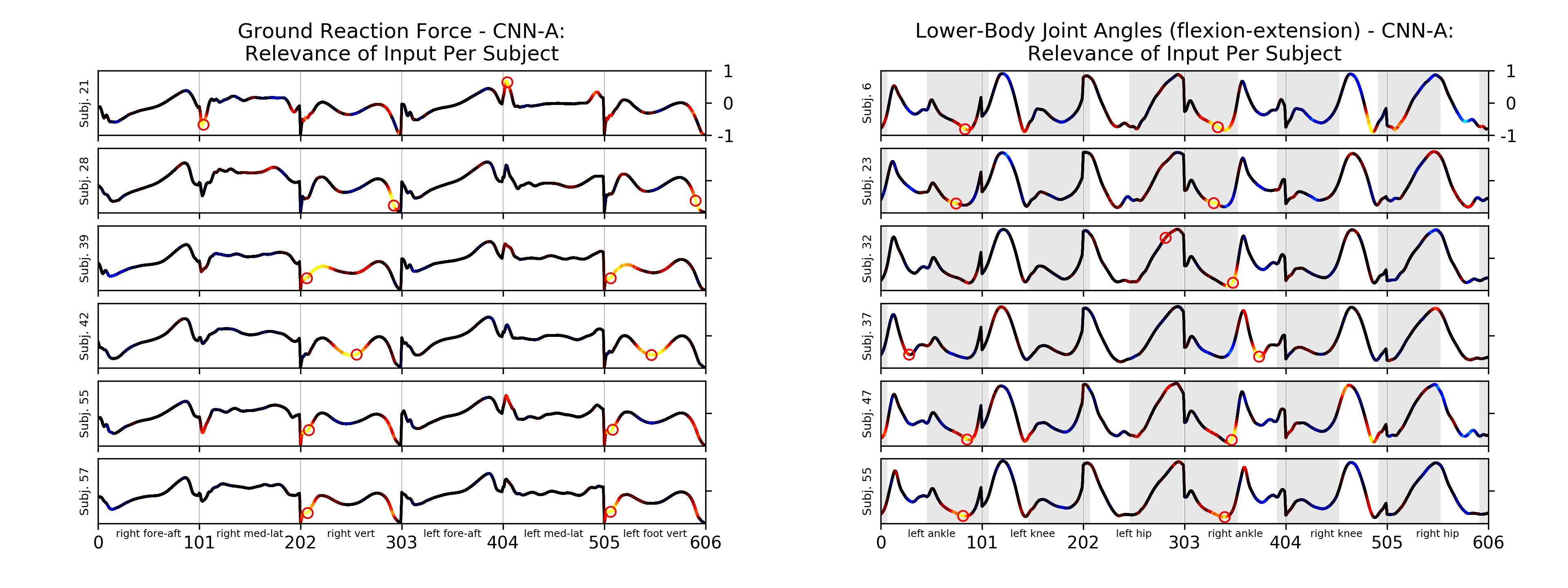}
    \caption{
    \emph{Left:} Mean Ground Reaction Force as a line plot, colour coded via input relevance values for the actual class for subject 21, 28, 39, 42, 55 and 57 using convolutional neural network CNN-A. The highest input relevance values per body side are highlighted by a red circle.
    \emph{Right:} Mean Lower-Body Joint Angles in the sagittal plane (flexion-extension) as line plot, colour coded via input relevance values for the actual class for subject 6, 23, 32, 37, 47 and 55 using convolutional neural network CNN-A. The highest input relevance values per body side are highlighted by a red circle. Shaded areas for the LBJAX highlight the time where the respective (left or right)
    foot is in contact with the ground.}
    \label{fig:cnna-subj}
    \end{center}
\end{figure}

As an example, Figure \ref{fig:overview} illustrates the unique gait signature from subject 6 by decomposing the input relevance values using LRP (see Figure \ref{fig:cnna-subj} and Supplementary Figures SF1 to SF4 for additional subject specific examples).
From the gait feature relevance, we can observe that the extension of the ankle during the terminal stance phase of the right and left leg and the flexion of the knee and hip during the initial contact of the right leg is unique to subject 6.
The kinetic data supports this finding, showing the highest input relevance values for the prediction of subject 6 for the vertical GRF during the terminal time window of the right and left stance phase.
On these grounds, LRP enables to discover the trial-individual gait signature from a certain individual. This individual signature can serve as indicator in clinical diagnoses and starting point for therapeutic interventions.
In our example (Figure \ref{fig:overview}), the terminal stance phase is unique to subject 6.
While this uniqueness might be interpreted as a reflection of a highly coordinated individual system, it could also indicate first relevant information about (forthcoming) complaints or impairments.
Clinicians and researchers are therefore capable to pick up the unique peculiarity during the terminal stance phase for an individualisation of therapeutic interventions, \eg by changing the strength of the responsible muscles for the extension of the ankle joint or reducing forces during the terminal stance phase during walking by shoes or insoles.

As discussed earlier, linear classification models have often been used for the classification of gait patterns \cite{schollhorn2004applications, phinyomark2018analysis, figueiredo2018automatic}.
However, other domains in machine learning have shown that highly non-linear DNNs are capable to outperform linear and kernel based methods
\cite{lecun2015deep, lecun2012efficient, krizhevsky2012imagenet, szegedy2015going, szegedy2017inception, yoon2014convolutional}.
In this study, we therefore investigate the applicability of state-of-the-art non-linear machine learning models such as DNNs for gait analysis and additionally compare different linear and non-linear machine learning methods in terms of prediction accuracy, model robustness and decomposition of input relevance values using the LRP technique (see Methods for detailed description).

\subsection*{Prediction Accuracy and Model Robustness}
The mean prediction accuracy for the subject-classification (on an out of sample set also denoted as test set) are summarised in Table \ref{tab:performance} (and Supplementary Table ST1).
The most striking result to emerge from this table is that most of the tested models were able to predict the correct class (subject) with a high accuracy above 95.4\% (ground reaction forces), 99.9\% (full-body joint angles) and 99.9\% (lower-body joint angles).

\begin{table}
    \begin{center}
    \begin{tabular}{r|rrrrr}
    \textbf{Model}	& \textbf{Ground}		& \textbf{Joint Angles}		& \textbf{Joint Angles}		&	\textbf{Joint Angles}	& \textbf{Joint Angles} 	\\
                    & \textbf{Reaction}		& \textbf{Full-Body}		& \textbf{Full-Body}		&	\textbf{Lower-Body}		& \textbf{Lower-Body} 		\\
                    & \textbf{Forces [\%]}	& \textbf{[\%]}				& \textbf{(flex.-ext.) [\%]}	&	\textbf{[\%]}			& \textbf{(flex.-ext.) [\%]} 	\\
    \hline
    Linear (SGD)	& 95.4 (1.7)			& \textbf{100.0 (0.0)}		& 96.3 (1.8)				& \textbf{100.0 (0.0)}		& 91.5 (2.2)				\\
    Linear (SVM)	& \textbf{100.0 (0.0)}	& \textbf{100.0 (0.0)}		& \textbf{99.7 (0.4)}		& \textbf{100.0 (0.0)}		& \textbf{99.8 (0.6)}		\\
    MLP (3, 64)		& 88.3 (3.7)			& 99.9 (0.3)				& 84.3 (3.3)				& 99.9 (0.3)				& 68.5 (8.3)				\\
    MLP (3, 256)	& 96.6 (0.8)			& \textbf{100.0 (0.0)}		& 95.6 (2.6)				& \textbf{100.0 (0.0)}		& 89.4 (4.3)				\\
    MLP (3, 1024)	& 98.8 (1.3)			& \textbf{100.0 (0.0)}		& 97.8 (1.0)				& \textbf{100.0 (0.0)}		& 96.5 (1.2)				\\
    CNN-A 			& 99.1 (0.8)			& \textbf{100.0 (0.0)}		& 95.6 (1.7)				& 99.9 (0.3)				& 92.0 (3.9)				\\
    CNN-C3			& 99.2 (0.6)			& \textbf{100.0 (0.0)}		& 97.7 (1.5)				& \textbf{100.0 (0.0)}		& 97.0 (1.3)				\\
    \end{tabular}
    \end{center}
    \caption{The prediction accuracy of the subject-classification task, reported as pairs of $mean~(standard~deviation)$ in percent.}
    \label{tab:performance}
\end{table}

The results in Table \ref{tab:performance} (and Supplementary Table ST1) are quite revealing in several aspects.
First, the highest prediction accuracy can be observed throughout the kinetic and kinematic variables for the linear support vector machines (Linear (SVM)).
Second, the linear one-layer fully-connected neural network using Stochastic Gradient Descent (Linear (SGD)), fully-connected neural network (multi layer perceptron (MLP)) using higher number of neurons (MLP (3, 256) and MLP (3, 1024)) and deep (convolutional) neural network architectures (CNN-A and CNN-C3) result in similar and throughout high prediction accuracies, while the prediction accuracy of fully-connected networks using a lower number of neurons (MLP (3, 64)) is decreased.
Third, surprisingly, even the linear neural network model architecture (Linear (SGD)) was able to predict the correct individual by quite high mean accuracies of 95.4\% (ground reaction forces), 100.0\% (full-body joint angles) or 100.0\% (lower-body joint angles).

While the prediction accuracy of the linear neural network model (Linear (SGD)) is comparable to the fully-connected and convolutional neural network architectures, the robustness of their predictions against noise on the testing data exhibits considerable differences.
As an example, Figure \ref{fig:perturbations} shows the progression of the mean prediction accuracy of the subject-classification for the stepwise increase of random noise perturbation on the test data.
That is, for a given input, we compute a random order by which the components of said input are perturbed, \eg with the addition of random gaussian noise to the current input component.
After each perturbation step (and executing the n-th perturbation for all test samples simultaneously), the prediction performance of a model is re-evaluated.
If a model is sensitive to random noise in the data it will react strongly to the ongoing perturbations.
Figure \ref{fig:perturbations} shows for each model the average test set prediction accuracy over 50 consecutive perturbation steps, averaged over 10 repetitions of the experiment.
It is apparent that the prediction accuracy of the Linear (SGD) model decreases rapidly after only a few perturbation steps, indicating low robustness and reliability of the model.
Variability is an inherent feature of human movements that occurs not only between but also within individuals. Therefore robust and reliable model predictions are a mandatory prerequisite for the development of automatic classification tools in clinical gait analysis.

Table \ref{tab:perturbations} (and Supplementary Table ST2)
summarises the robustness of the model predictions and presents the mean area over perturbation curve (AOPC)\cite{SamTNNLS17} of the stepwise increase of random noise perturbation over multiple repetitions of the perturbation runs on the test data.
High AOPC values computed with above strategy of random perturbations corresponds to high sensitivity to noise.
From the results presented in
Table \ref{tab:perturbations} (and Supplementary Table ST2)
, it is apparent that the stepwise increase of noise on the test data lead to an abrupt decrease in the prediction accuracy of the linear neural network. Furthermore, closer inspection shows that fully-connected models based on a higher number of neurons are more robust against noise perturbations on the test data than models based on lower numbers of neurons.
Even though the prediction problem at hand seems simple enough such that linear predictors perform excellently (\ie the relationship between the input variables and the prediction target is linearly separable) and the performance-wise gain from non-linear and networks is minimal, the deeper architectures bring to the table considerably more robust predictors, which is especially valuable in application settings like gait analysis, where variability is an important factor to consider.

\begin{table}
    \begin{center}
    \begin{tabular}{r|rrrrrrr}
    \textbf{Model /}& \textbf{Gaussian}		& \textbf{Gaussian}			& \textbf{Gaussian}			&	\textbf{Salt$^-$}		& \textbf{Pepper} 	 	& \textbf{Salt$^+$}	& \textbf{Shot}				\\
    \textbf{Noise  }& \textbf{Random}		& \textbf{Random}			& \textbf{Random}			&	\textbf{Random}			& \textbf{Random} 	 	& \textbf{Random}		& \textbf{Random} 		\\
                    & \textbf{$\sigma=0.5$}	& \textbf{$\sigma=1.0$}		& \textbf{$\sigma=2.0$}		&							&  	\\
    \hline
    Linear (SGD)	& 40.5 (6.0)			& 45.6 (3.0)				& 47.6 (1.8)				& 46.1 (2.7)				& 37.1 (7.0)			&	46.4 (3.1)			&	46.5 (2.9)			\\
    Linear (SVM)	& \textbf{4.0 (3.9)}	& 18.6 (7.0)				& 35.4 (4.7)				& 31.3 (7.7)				& \textbf{2.5 (2.4)}	&	\textbf{32.8 (8.7)}	&	\textbf{32.4 (8.6)}	\\
    MLP (3, 64)		& 16.2 (14.0)			& 27.6 (12.6)				& 37.8 (9.0)				& 43.5 (7.1)				& 14.4 (13.6)			&	43.3 (7.4)			&	43.3 (7.4)			\\
    MLP (3, 256)	& 8.9 (9.7)				& 20.8 (10.9)				& 34.9 (8.4)				& 41.7 (6.8)				& 7.7 (9.5)				&	41.2 (7.4)			&	41.1 (7.5)			\\
    MLP (3, 1024)	& 5.4 (7.0)				& \textbf{16.7 (9.4)}		& \textbf{32.4 (7.8)}		& 40.0 (6.6)				& 4.3 (6.1)				&	39.4 (7.2)			& 	39.4 (7.2)			\\
    CNN-A 			& 6.1 (6.8)				& 18.2 (9.4)				& 33.1 (8.0)				& 36.9 (6.6)				& 8.8 (8.5)				&	37.4 (7.5)			&	37.4 (7.6)			\\
    CNN-C3			& 12.2 (8.6)			& 27.5 (9.4)				& 38.0 (7.0)				& \textbf{31.0 (8.5)}		& 11.3 (8.6)			&	37.9 (8.2)			&	38.1 (8.1)			\\
    \end{tabular}
    \end{center}
    \caption{The area over perturbation curve of the subject-classification of ground reaction forces for different noise perturbation runs (AOPC)\cite{SamTNNLS17}.
    Values are reported as pairs of $mean~(standard~deviation)$.
    Smaller values correspond to higher robustness.}
    \label{tab:perturbations}
\end{table}

\subsection*{Interpreting and Understanding Model Predictions using Layer-Wise Relevance Propagation}
As an example Figure \ref{fig:cnna-subj} shows relevance attributions for predictions based on ground reaction force (Figure \ref{fig:cnna-subj}  (left)) and the lower-body joint angles (Figure \ref{fig:cnna-subj} (right)) of a certain individual based on his or her gait patterns.

\begin{figure}
    \begin{center}
    \includegraphics[width=0.6\textwidth]{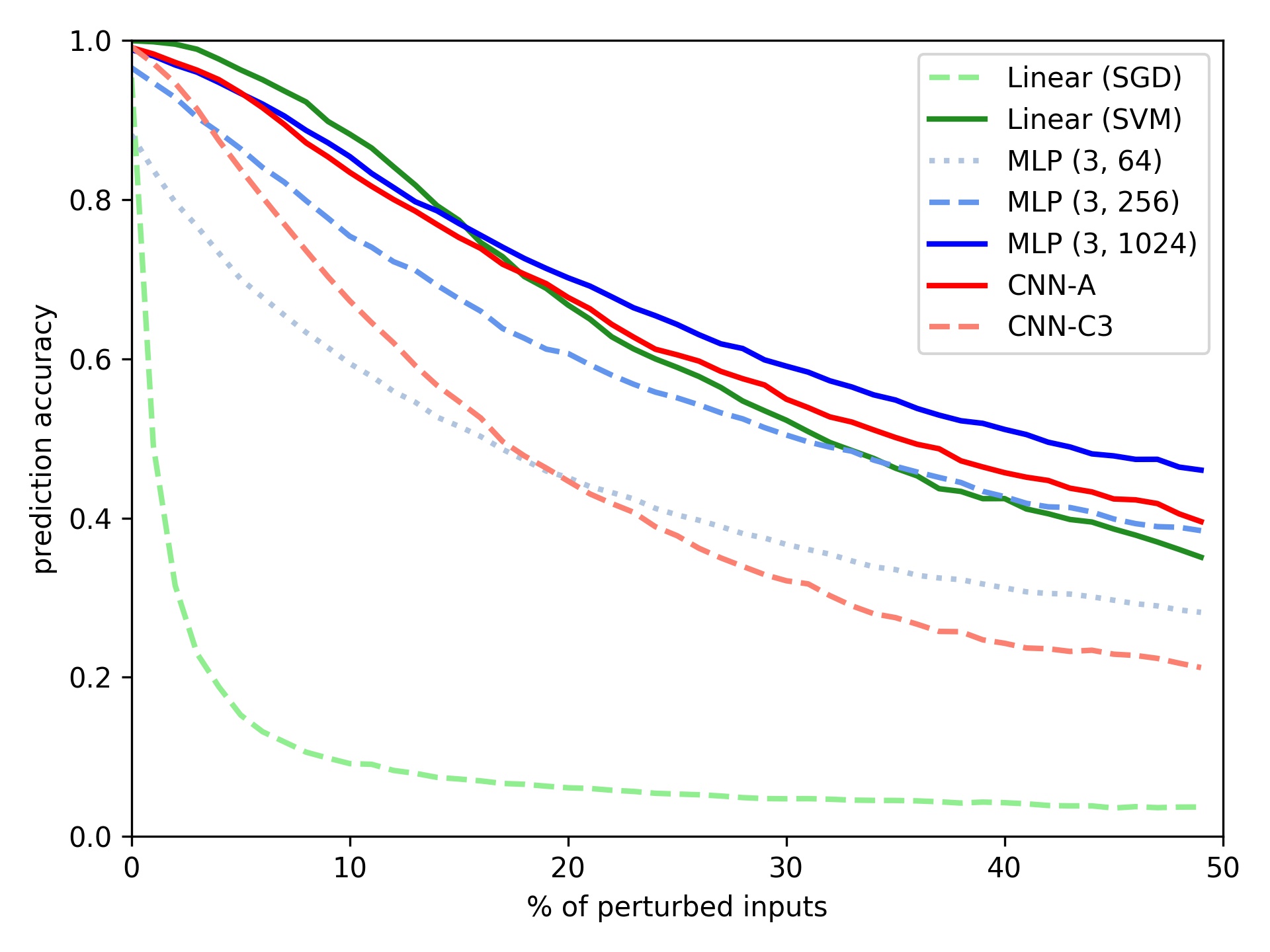}
    \caption{Progression of the mean prediction accuracy of the subject-classification of ground reaction forces for stepwise random perturbation using gaussian noise with $\sigma=1$.}
    \label{fig:perturbations}
    \end{center}
\end{figure}

The input relevance contributions point out which gait characteristics were most relevant for the identification of an individual and thereby reveal the unique gait signature of a certain subject.
The comparison of input relevance values from different subjects indicates that individuals were classified by both, different gait characteristics and differing magnitudes or shapes of the same gait characteristic.
For example, the highest input relevance values for the prediction of subject 21 (Figure \ref{fig:cnna-subj} (left)) can be observed in the medial/lateral ground reaction force at approximately 10 \% of the gait stride and subject 28 (Figure \ref{fig:cnna-subj} (left)) in the vertical ground reaction force at approximately 90 \% of the gait stride.
While the highest input relevance values for the prediction of subject 55 (Figure \ref{fig:cnna-subj} (left)) and subject 57 (Figure \ref{fig:cnna-subj} (left)) can be both observed in the vertical ground reaction force at approximately 10 \% of the stance phase.
It is further interesting that among all predicted LBJAX curves shown in Figure \ref{fig:cnna-subj} (right), subject 37 is the only one that is identified dominantly by gait characteristics during the swing phase (and not the stance phase).
The model has associated the pronounced flexion of the right (left) ankle joint during the swing phase unique to subject 37.

\begin{figure}
    \begin{center}
    \includegraphics[width=\textwidth]{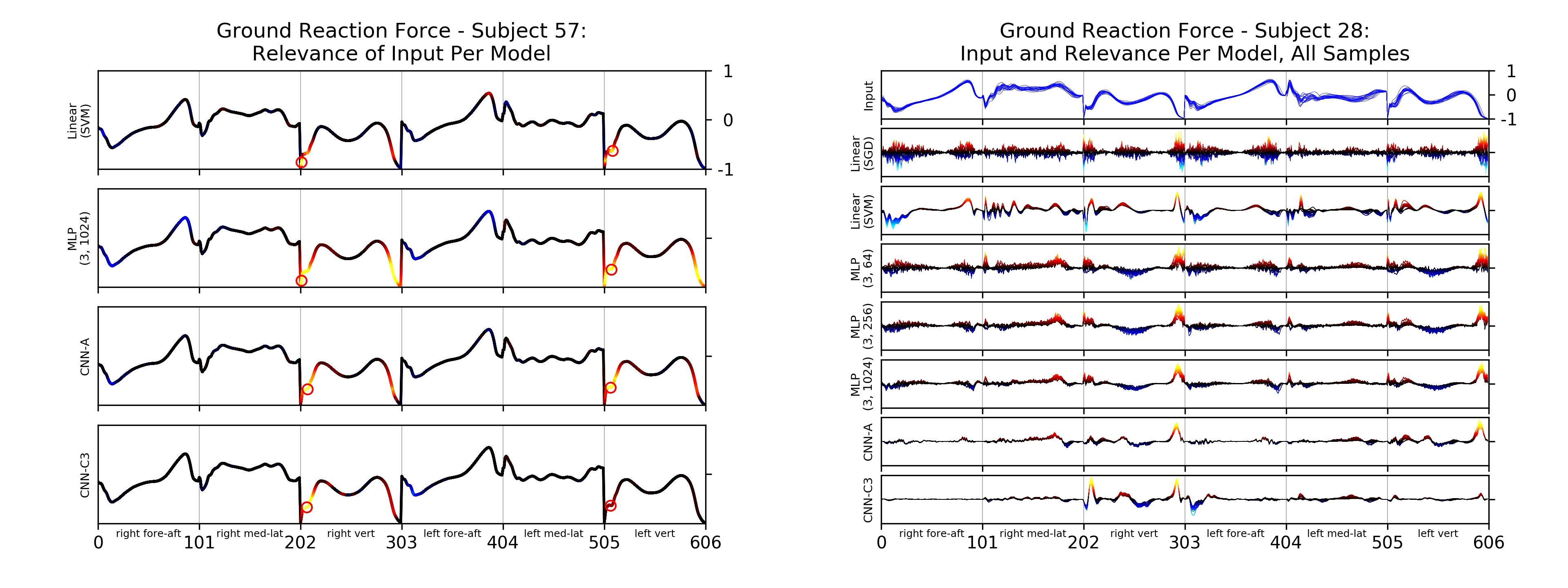}
    \caption{
    \emph{Left:} Mean Ground Reaction Force as line plot, colour coded via input relevance for the actual class of different models using artificial neural networks and the linear SVM model from subject 57. The highest input relevance values per body side are highlighted by a red circle.
    \emph{Right:} Input relevance as colour coded line plots for the predicted class of different models using artificial neural networks and linear models of ground reaction force of the 20 gait trials from subject 28.}
    \label{fig:grf}
    \end{center}
\end{figure}

In the vast majority of cases, the input relevance values for a certain class (individual) are comparable between the different model architectures (Figure \ref{fig:grf} (left)), \ie all models pick up on similar features, which are characteristic to the individual subject in general.
It is rather apparent from the results that most artificial neural networks and the linear SVM are using a number of different gait characteristics for their predictions, albeit the vast majority of the inputs seems to be irrelevant to the model’s decision (see Figure \ref{fig:overview}; the tight interval ($R \approx 0$) projected onto black), which are largely based on individual details in a subject’s movement patterns.
This explanatory relevance feedback indicates that (non-linear) machine learning methods such as (deep) artificial neural networks are not arbitrarily picking up single randomly distributed input values, but rather learn certain dynamically meaningful features that can be related to functional gait characteristics and thus making them applicable in clinical gait analysis.

When comparing input relevance values from different architectures of artificial neural networks and the SVM strikingly all model architectures (except the CNN-C3) show that not a single gait characteristic (certain variable at a certain time window of the gait cycle) is relevant for the identification of a certain individual, but rather complex combinations thereof.

Even more interesting is the observation that input relevance value attributions appear to be similar between right and the left body side variables (as an example take Figure \ref{fig:grf} (left): Linear (SVM), MLP (3, 1024) and CNN-A).
This indicates the importance of symmetries/asymmetries between left and right body side variables for the examination of human gait.
Note however that the models all learned without any information about the physiological meaning of the used variables.
While the input relevance attributions are throughout comparable for all evaluated architectures -- \ie there is, given a subject, a non-empty set of features recognized as characteristic features to the individual by all models -- some models have learned to ground their predictions on supplementary gait characteristics (\eg Figure \ref{fig:grf} (left): CNN-C3 vs CNN-A).

We can further observe that the use of multiple gait characteristics for prediction can be associated with a model’s robustness to random perturbations of the input.
The observed robustness of the evaluated models is also reflected by the reliability of the attributed input relevance values:
Over the relevance values of each input component and subject (\ie 20 relevance values each), the coefficient of variation\cite{winter1984kinematic} was computed in order to prove their consistency over several trials and cross validation splits.
The coefficient of variation represents the (root mean square) normalized band of standard deviation around the relevance signal of an input variable, where low values correspond to high reliability/stability of relevance attributions to a observed feature between samples and data splits, and thus to the model's ability to generalize.
A high coefficient of variation indicates that a model overfits on its respective training split population.
As an example, Figure \ref{fig:grf} (right) shows the input relevance values for the actual class prediction for the ground reaction forces.
Qualitatively, the highest deviations of input relevance values between trials appeared in the prediction of the linear neural network model (Linear (SGD)), which also is most sensitive to even minute noise added to the test data (Figure \ref{fig:perturbations} and Table \ref{tab:perturbations}).
It becomes apparent from Figure~\ref{fig:grf}~(right) that the variance of input relevance is decreasing in fully-connected neural network architectures composed of increasing numbers of neurons.
However, the lowest variance of the input relevance values can be observed in the relevance decomposition of the predictions from the Linear (SVM) and convolutional neural network architectures, which we attribute for the former to the complexity of the regularised training regime and complexity of the model itself for the latter.

Table \ref{tab:coefficientofvariation} (Supplementary Table ST3) summarises the mean coefficient of variation for the input relevance values for the subject-classification over all subjects, expressing that decreasing variance(increasing reliability) goes along with increasing model complexity, and also model robustness when compared to Figure \ref{fig:perturbations} and Table \ref{tab:perturbations}.
Hence, the lowest reliability is present for the Linear (SGD) model, while reliability is increasing in fully-connected model architectures composed of increasing numbers of layers and neurons and discloses the highest reliability for the convolutional neural networks (CNN-A and CNN-C3).
Interestingly, the reliability of the input relevance values from linear support vector machines (Linear (SVM)) are as well on a high level and comparable to convolutional neural networks.

\begin{table}
    \begin{center}
    \begin{tabular}{r|rrrrr}
    \textbf{Model}	& \textbf{Ground}		& \textbf{Joint Angles}		& \textbf{Joint Angles}		&	\textbf{Joint Angles}	& \textbf{Joint Angles} 	\\
                    & \textbf{Reaction}		& \textbf{Full-Body}		& \textbf{Full-Body}		&	\textbf{Lower-Body}		& \textbf{Lower-Body} 		\\
                    & \textbf{Forces}		& 							& \textbf{(flex.-ext.)}			&							& \textbf{(flex.-ext.)}	 	\\
    \hline
    Linear (SGD)	& 4.31 (0.25)			& 4.27 (0.08)				& 3.93 (0.12)				& 4.16 (0.10)				& 3.86 (0.15)				\\
    Linear (SVM)	& 0.31 (0.08)			& 0.56 (0.10)				& \textbf{0.31 (0.07)}		& \textbf{0.48 (0.09)}		& \textbf{0.26 (0.05)}		\\
    MLP (3, 64)		& 1.31 (0.34)			& 2.73 (0.15)				& 1.58 (0.17)				& 2.37 (0.16)				& 1.41 (0.23)				\\
    MLP (3, 256)	& 0.84 (0.18)			& 2.21 (0.13)				& 1.05 (0.10)				& 1.86 (0.13)				& 0.85 (0.12)				\\
    MLP (3, 1024)	& 0.63 (0.12)			& 1.85 (0.10)				& 0.77 (0.08)				& 1.50 (0.10)				& 0.61 (0.09)				\\
    CNN-A 			& \textbf{0.30 (0.08)}	& 0.56 (0.09)				& 0.35 (0.06)				& 0.49 (0.08)				& 0.32 (0.05)				\\
    CNN-C3			& 0.35 (0.08)			& \textbf{0.50 (0.08)} 		& 0.43 (0.08)				& \textbf{0.48 (0.08)}		& 0.44 (0.08)				\\
    \end{tabular}
    \end{center}
    \caption{The coefficient of variation of the input relevance values of the subject-classification, reported in pairs of $mean~(standard~deviation)$ over all subjects.}
    \label{tab:coefficientofvariation}
\end{table}

\section*{Discussion}
The present results verified the uniqueness of characteristics for individual gait patterns based on kinematic and kinetic variables.
By decomposing the prediction of machine learning methods such as (deep) artificial neural networks back to the input variables (time-continuous ground reaction forces and full-body joint angles), the LRP technique demonstrated which gait variables were most relevant for the characterisation of gait patterns from a certain individual.
By measuring the contribution of each input variable to the prediction of (deep) artificial neural networks, the present paper describes a procedure that enables to understand and interpret the predictions of non-linear machine learning methods in (biomechanical) gait analysis.
LRP thereby outlines the first general framework that facilitates to overcome the inherent black box problem of non-linear machine learning methods and makes them applicable in clinical gait analysis.
In the context of personalised medicine, the determination of characteristics that are specific for gait patterns of a certain individual facilitates to support clinicians and researchers in the individualisation of their analyses, diagnoses and interventions.

The individual nature of human gait patterns was quantified using different linear and non-linear machine learning methods.
The present results support previous studies on the individuality of human movements \cite{schollhorn2002identification, horst2017one, janssen2008recognition, janssen2011diagnosing} and provide evidence for gait characteristics that are unique to an individual and can be clearly differentiated from gait patterns of other individuals.
Most of the artificial neural network architectures classified gait patterns almost error-free to the corresponding individual and achieved very high prediction accuracies that are suitable for clinical applications.
However, advantages for more sophisticated model architectures (like fully-connected model using a higher numbers of neurons or deep convolutional neural networks) can be observed in higher prediction accuracies (Table \ref{tab:performance} and Supplementary Table ST1) and even more significant in the higher robustness of the model predictions against noise perturbations on the test data (Figure \ref{fig:perturbations}, Table \ref{tab:perturbations} and Supplementary Table ST2).
Because variability within individuals \cite{horst2016daily, horst2017intra} as well as variability due to differences between individuals \cite{schollhorn2002identification, horst2017one}, genders \cite{begg2005machine} and ages \cite{eskofier2013marker}  is an inherent feature of human motor control, prediction accuracy and model robustness are both essential for the development of reliable clinical applications using machine learning.
Consequently, the present results suggest high potential  of state-of-the-art non-linear methods such as DNNs compared to linear methods.

One of the issues that emerges from the evidence that gait patterns are unique to an individual, is the demand to evaluate clinical approaches for diagnoses and therapy that consider individual needs\cite{schollhorn2002identification, horst2017one}.
However, previous studies could not address how an individualisation of diagnoses and therapy could be obtained.
By measuring the contribution of each input variable to the prediction of machine learning methods, the LRP method enables one for the first time to describe qualitatively why a certain individual could be identified based on his/her gait patterns.
The LRP technique provides the possibility to comprehend what a model has learned and to interpret the input relevance values as representation for a certain class (individual).
In the context of personalised gait analysis (medicine) that means, the decomposition of input relevance values and their dynamics describe what input variables are most relevant for the identification of a certain individual and thereby indicate which input variables are the most characteristic ones for the gait patterns of a certain individual (Figure \ref{fig:overview}).

On these grounds, the input relevance values enable clinicians and researchers to determine the unique gait signature of a certain individual based on single trials and adjust their analyses, diagnoses and interventions to the specific needs of this individual (Figure~\ref{fig:overview}, Figure~\ref{fig:cnna-subj}).

In addition, explaining the model predictions provide interesting insights into the analysis of gait patterns.
The input relevance values highlight that in most cases not a single gait characteristic (specific value or shape of a certain variable at a certain time of the gait cycle) is relevant for the identification of a certain individual.
It is rather apparent that most artificial neural networks architectures look for the shape of different variables as well as their interaction at the same time window or at different time windows of the gait cycle.
Similar results have been found on photographic image data \cite{BachPLOS15,LapAMFG17}.
Interestingly, the prediction of most artificial neural network architectures (except CNN-C3) trace to input relevance values that are similar between right and left body side variables at the same time.
That means a certain variable at a certain time window of the gait cycle of the right and the left body side is relevant for the prediction of the models (Figure \ref{fig:grf} (left)), indicating importance of symmetries and asymmetries between right and left body movements for the identification of individuals and probably the examination of human gait in general.

The input relevance values support that machine learning approaches like artificial neural networks are able to consider several variables at various time points of the gait cycle for their predictions.
In comparison to most conventional approaches of gait analysis that are based on single pre-selected variables, machine learning approaches seem to be promising to represent the multi-dimensional associations of human locomotion and their connections to functional and neurological disease \cite{schollhorn2002identification, federolf2012holistic,schollhorn2004applications}.

The present results demonstrate in the vast majority of cases that the input relevance values are similar between different model architectures (Figure \ref{fig:grf}).
That means all models pick up similar features for the classification of gait patterns, which are characteristic to the individual subject in general.
However, the LRP technique enables to identify the strategy of a certain model to classify a class (individual gait patterns) and to compare strategies between different model architectures \cite{LapAMFG17}.
For an implementation of machine learning in clinical diagnoses and therapeutic interventions, for example in terms of an automatic classification of gait disorders or (neurological) disease \cite{figueiredo2018automatic, zeng2016parkinson, alaqtash2011automatic}, the understanding about their decisions and decision-making seems to be inevitable.
Since the lack of transparency has so far been a major drawback of preceding applications of machine learning, \eg in medical applications (like gait analysis), further research on explaining, understanding and interpreting machine learning predictions should get attention.

Here, the decomposition of input variable relevance values using the LRP was consistent over multiple test trials and cross-validation splits.
Taken together, the results demonstrate the suitability of the proposed method for the explanation of machine learning predictions in clinical (biomechanical) gait analysis.
However, higher reliability of input relevance values between test trials and cross-validation splits indicate advantages for deep (convolutional) neural networks architectures.
These findings are in agreement with those observed in earlier studies on text \cite{arras2017relevant} or image \cite{LapCVPR16} classification that indicated more robust and traceable class representations in deep (convolutional) neural networks.

In conclusion, the present findings underline that methods enabling to understand and interpret the predictions of machine learning, like the LRP, are highly promising for the application and implementation of machine learning in gait analysis.
Due to the above discussed advantages of non-linear machine learning methods such as DNNs for the analysis of human gait \cite{chau2001review1, chau2001review2, schollhorn2004applications, phinyomark2018analysis}, the understanding and interpreting of machine learning predictions is essential in order to overcome one of their major drawbacks (the lack of transparency) \cite{chau2001review1, chau2001review2, wolf2006automated}.
Using the testbed of uniqueness of individual gait patterns, the present study proposed a general framework for the understanding and interpretation of non-linear machine learning methods in gait analysis thus providing a solid basis for future studies in biomechanical analysis and clinical diagnosis.

\section*{Methods}
\label{sec:methods}

\subsection*{Subjects and ethics statement}
Fifty-seven physically active subjects ($29$ female, $28$ male; $23.1 \pm 2.7$ years; $1.74 \pm 0.10$ m; $67.9 \pm 11.3$ kg) without gait pathology and free of lower extremity injuries participated in the study.
The study was carried out according to the Declaration of Helsinki at the Johannes Gutenberg-University in Mainz (Germany). All subjects were informed about the experimental protocol and provided their informed written consent to participate in the study. Subjects appearing in the figures provided informed written consent to the publication of identifying images and videos in an online open-access publication. The approval from the ethical committee of the medical association Rhineland-Palatinate in Mainz (Germany) was received.

\subsection*{Experimental protocol and data acquisition}
\begin{figure}
    \begin{center}
    \includegraphics[width=0.5\textwidth]{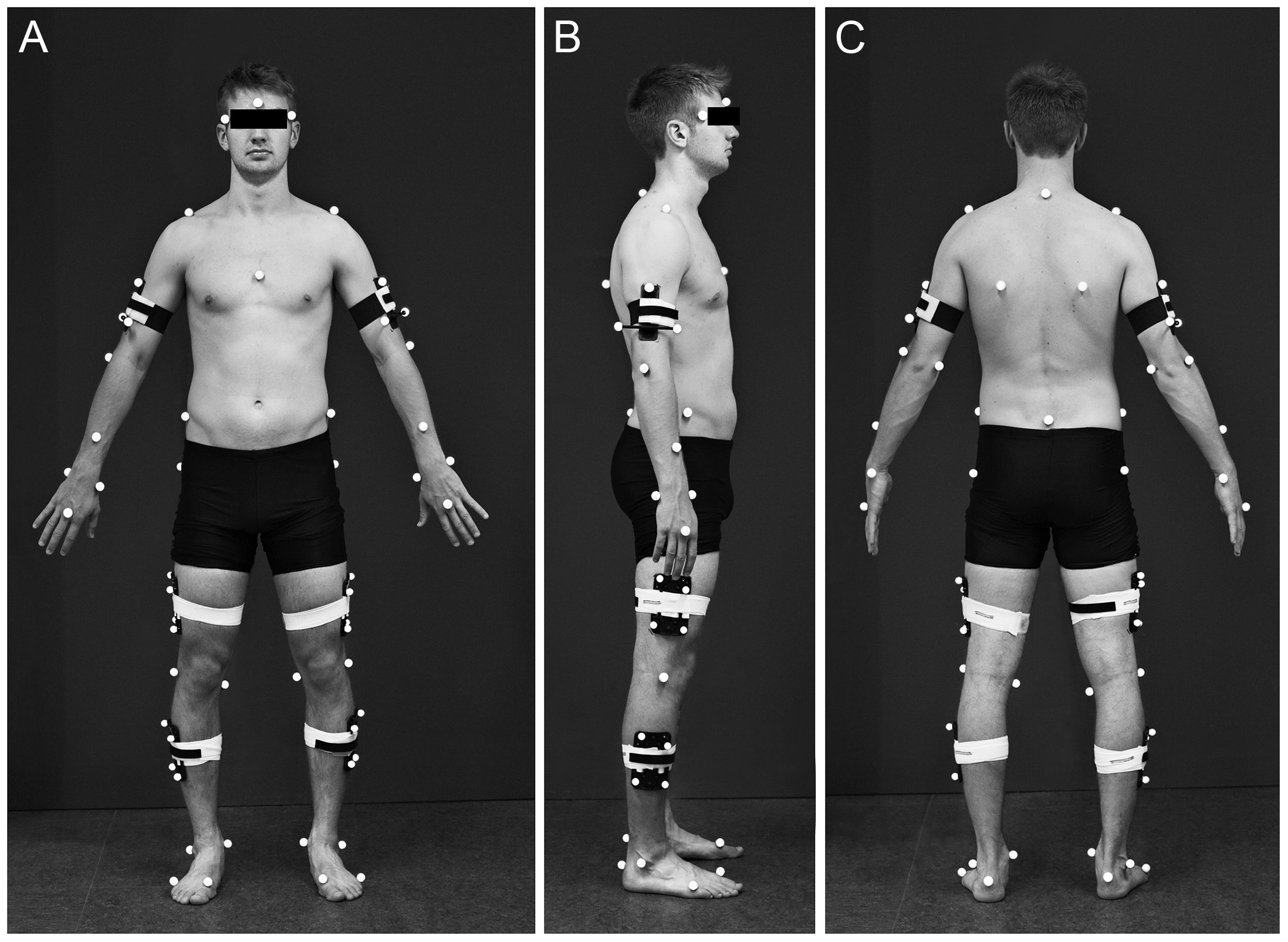}
    \caption{Full body marker set in (A) anterior (B) right lateral (C) posterior view.
    The markers were placed at os frontale glabella, 7th cervical vertebrae, sternum jugular notch, sacrum (mid-point between left and right posterior superior iliac spine) and bilaterally at greater wing of sphenoid bone, acromion, scapula inferior angle, humerus lateral epicondyle, humerus medial epicondyle, forearm, radius styloid process, ulna styloid process, head of 3rd metacarpal, iliac crest tubercle, femur greater trochanter, femur lateral epicondyle, femur medial epicondyle, fibula apex of lateral malleolus, tibia apex of medial malleolus, posterior surface of calcaneus, head of 1st metatarsus, head of 5th metatarsus and clusters with four markers each at the thigh and shank and clusters of three markers each at the humerus.}
    \label{fig:markerplacement}
    \end{center}
\end{figure}

The subjects performed 20 gait trials in a single assessment session, while they did not undergo any intervention.
For each trial upper- and lower-body joint angles as well as ground reaction forces were measured, while the subjects walked on a 10 m path.
The subjects were instructed to walk barefoot at a self-selected speed.
Kinematic data were recorded using a full-body marker set consisting of 62 retro reflective markers placed on anatomical landmarks (Figure \ref{fig:markerplacement}).
Ten Oqus 310 infrared cameras (Qualisys AB, Sweden) captured the three-dimensional marker trajectories at a sampling frequency of 250 Hz.

The three-dimensional ground reaction forces were recorded by two Kistler force plates (Type 9287CA) (Kistler, Switzerland) at a frequency of 1000 Hz.
The recording was managed in time-synchronization by the Qualisys Track Manager 2.7 (Qualisys AB, Sweden).
Two experienced assessors attached the markers and conducted the analysis. Every subject was analysed by the same assessor only.
The laboratory environment was kept constant during the investigation.

Before the data acquisition, each subject performed 20 test trials to get accustomed to the experimental setup and to assign a starting point for a walk over the force plates.
This procedure is described to minimize the impact of targeting on the force plates on the observed gait variables \cite{sanderson1993effects,wearing2000effect}.
Additionally, the participants were instructed to watch a neutral symbol (smiley) on the opposing wall of the laboratory to direct their attention away from targeting on the force plates and ensure a natural walk with an upright body position.

\subsection*{Data processing}
The gait analysis was conducted for one gait stride per trial.
The stride was defined from right foot heel strike to left foot toe off and was determined using a vertical ground reaction force threshold of 10 N.
The three-dimensional marker trajectories and ground reaction forces were filtered using a second order Butterworth bidirectional low-pass filter at a cut off frequency of 12 Hz and 50 Hz, respectively.
The ground reaction force data were normalized to the body weight.
The computation of the upper- and lower-body joint angles was conducted by Visual3D Standard v4.86.0 (C-Motion, USA) for elbow, shoulder, spine, hip, knee and ankle in sagittal, transversal and coronal plane.

Further data processing was executed by a self-written script within the software Matlab 2016a (MathWorks, USA).
Each variable time course was normalized to 101 data points, z-transformed and scaled to a range of -1 to 1\cite{hsu2003practical}.
The z-transformation was executed for kinematic variables for each trial separately and for kinetic variables for all trials.
The scaling was carried out in order to prevent numerical difficulties during the calculation of the artificial neural networks \cite{hsu2003practical} and to ensure an equal contribution of all variables to the classification rates and thereby avoid that variables in greater numeric ranges dominate those in smaller numeric ranges\cite{hsu2003practical}.
Scaling is a common procedure for data processing in advance for the classification of gait data\cite{schollhorn2002identification,figueiredo2018automatic}.

\subsection*{Data Analysis}
The data analysis was conducted within the software frameworks of Matlab 2016a (MathWorks, USA) and Python 2.7 (Python Software Foundation, USA).
The ability to distinguish gait patterns of one subject from gait patterns of other subjects was investigated in a multi-class classification (subject-classification) using the data from 57 subjects.
The classification of gait patterns, based on time-continuous kinetic and kinematic data, was carried out by supervised machine learning models using support vector machines (SVM)\cite{boser1992training,cortes1995support,muller2001introduction,scholkopf2002learning} and artificial neural networks (ANN)\cite{wolf2006automated,lecun2015deep,lecun2012efficient}.
While fully-connected ANNs such as multi layer perceptions (MLP) and SVM represent established models for the classification of gait patterns based on joined input vectors of time-continuous kinematic and kinetic data\cite{schollhorn2004applications}, convolutional (deep) artificial neural networks (DNN) have not yet been applied for the biomechanical analysis of human movements.
Because DNNs showed superior prediction accuracies in domains like image\cite{krizhevsky2012imagenet,szegedy2015going,szegedy2017inception} or speech recognition\cite{yoon2014convolutional}, they seem to be promising for the given classification of human gait patterns.

In the present paper, an SVM and different architectures of fully-connected and convolutional artificial neural networks were compared in terms of prediction accuracy, model robustness to noise on the test data and decomposition of input relevance values.

As a simple baseline, two linear classification models were implemented.
A one-layer fully-connected neural network (Linear (SGD)) and a SVM using a linear kernel function (Linear (SVM)).
Among the considered fully-connected artificial neural network architectures were all combinations $L \times H$ with $L \in [2,3]$ describing the number of layers and $H \in [2^6,\dots,2^{10}]$ describing the number of neurons per hidden layer.
For the convolutional artificial neural network architectures, the number of convolutional layers is $C \in [1,2,3]$ with the number of hidden neurons depending on the number of channels in the data as well as the stride and shape of the learned convolutional filters.
All convolutional neural network architectures are topped off with one linear layer, connecting all neurons of the highest convolutional layer to the number of classes of the prediction problem.
Major architectural differences between the evaluated convolutional neural network architectures were the sizes of the input layer filters ( $3 \times 3$ and $6 \times 6$ as well as $C \times 3$ , $C \times 6$  and $C \times C$ ) spanning different amounts of neighbouring channels and time windows  in the input samples.

All artificial neural networks using hidden layers (i.e. all architectures except the linear classifier Linear (SGD)) have ReLU-nonlinearities after each linear / convolutional layer as activation functions for the hidden neurons and a SoftMax activation function for the output layer.
Both linear and the fully-connected classifiers receive as input the channel~$\times$~time samples as row-concatenated channel~$\cdot$~time dimensional vectors.
The convolutional models directly operate on the channel~$\times$~time shaped samples.
For a detailed description of all evaluated model architectures, see the Supplement (Supplementary Tables ST4-ST9).

With the exception of the linear support vector machines predictor, all models have been trained as n-way classifiers using Stochastic Gradient Descent (SGD) Optimization\cite{lecun2012efficient} for up to $3\cdot10^4$ iterations of mini batches of 5 randomly selected training samples and an initial learning rate of $5e^{-3}$.
The learning rate is gradually lowered to $1e^{-3}$ and then $5e^{-4}$ after every $10^4$ training iterations.
Model weights are initialized with random values drawn from normal distributions with $\mu=0$ and $\sigma=m^{-\frac{1}{2}}$, where $m$ is the number of inputs to each output neuron of the layer\cite{lecun2012efficient}.
The linear support vector machine model has been using regularized quadratic optimization.

For SVM, the multi-class linear support vector classifier of the scikit learn Toolbox for python~\cite{pedregosa2011scikit} was used with a regularisation parameter $C = 0.1$.

Prediction accuracies were reported over a ten-fold cross validation configuration, where eight parts of the data are used for training, one part is used as a validation set and the remaining part is reserved for testing.
With on average 912 samples per split being reserved for training, a (neural network) model passes the training set up to 164 (= 30000 iterations $\cdot$ 5 samples per batch / 912 training samples on average) times and each training stage tied to a given learning rate may be terminated prematurely if the model performance has converged on the validation hold out set to avoid overfitting on the training data.
For subject-classification, the 20 samples per subject are uniformly distributed across all data partitions at random.

\subsection*{Layer-wise Relevance Propagation}
One of the main reasons for the wide-spread use of linear models in the (meta) sciences is the inherent transparency of the prediction function.
Given a set of learned model parameters $\lbrace \w, \b \rbrace$ where $\w$ is a weight vector matching the dimensionality of the input data and a bias term $\b$, the (multi-class) prediction function for an arbitrary input $\x$ evaluates for class $c$ as
\begin{align}
f_c(\x)~=~\w^T\x~+~b_c~=~\sum_i~w_{ic}x_i~+~b_c~.
\end{align}
It is apparent, that component $i$ of the given input $\x$ contributes to the evaluation of $f_c$ together with the learned parameters as the quantity $w_{ic}x_i$.
Each decision made by a linear model is therefore transparent, while complex non-linear models are generally considered black box classifiers.

A technique called Layer-Wise Relevance Propagation (LRP)\cite{BachPLOS15} has generalized the explanation of linear models for non-linear models such as deep (convolutional) artificial neural networks and arbitrary pipelines of pre-processing steps and nonlinear predictions.
As a principled and general approach, LRP decomposes the output of a given decision function $f_c$ for an input $\x$ and attributes ``relevance scores'' $R_i$ to all components $i$ of $\x$,
such that $f_c(\x) = \sum_i R_i$.
Similarly to how the prediction of a linear model can be ``explained'' LRP starts at the model output by selecting a class output $c$ of interest, initiating $f_c(\x) = R_j$ (and selecting 0 for all other model outputs) as the initial output neuron relevance value.
Note that for two-class problems, there is often one shared model output, with the predicted class being determined by the sign of the prediction.
Here, $f(\x)=R_j$ initially.

The method can best be described by considering a single output neuron $j$ anywhere within the model.
That neuron receives a quantity of relevance $R_j$ from upper layer neurons (or is initiated with that value in case of a model output neuron, and redistributes that quantity to its immediate input neurons $i$ , in proportion to the contribution of the inputs $i$ to the activation of  $j$ in the forward pass:
\begin{align}
R_{i \leftarrow j} = \frac{z_{ij}}{z_j}R_j~.
\label{eq:lrpdecomp}
\end{align}
Here, $z_{ij}$ is a  quantity measuring the contribution of the input neuron $i$  to the output neuron $j$ and $z_j$ is the aggregation thereof.
This decomposition approach follows the semantic that the output neuron $j$ holds a certain amount of relevance, due to its activation in the forward pass and its influence to consecutive layers and finally the model output.
This relevance is then distributed across the neuron’s inputs in proportion of each input’s contribution to the activation of neuron:
If a neuron $i$ contributes as $z_{ij}$  towards the overall trend $z_j$ , it shall receive a positively weighted fraction of $R_j$.
If fires against the overall trend, \eg the amount or relevance attributed to it will be weighted negatively.

Usually, the layers of an ANN model implement an affine transformation function $x_j = \left(\sum_i x_iw_{ij} \right) +~b_j$ or a (component-wise) non-linearity $x_j = \sigma(x_i)$.
In the former case, we then have $z_{ij}=x_iw_{ij}$ , for example, and $z_j$ is the output activation $x_j$ .
In the latter case, $z_{ij} = \delta_{ij}\sigma(x_i)$ where $\delta_{ij}$ is the Kronecker delta, since there is no mixing between inputs and outputs of different subscripts.

The relevance score $R_i$ at input neuron $i$  is then obtained by pooling all incoming relevance values $R_{i \leftarrow j}$  from the output neurons to which $i$ contributes in the forward pass:
\begin{align}
R_i = \sum_j R_{i \leftarrow j}~.
\label{eq:lrppool}
\end{align}
Together, both above relevance decomposition and pooling steps ensure a local relevance conservation property, \ie $\sum_i R_i = \sum_j R_j$  and thus $\sum_i R_i = f(\x)$ for all layers of the model.
In case of a component-wise operating non-linear activation, \eg a ReLU ($\forall_{i=j}: \x_j = \max(0,x_i)$) or Tanh ($\forall_{i=j}: x_j = \tanh(x_i)$), then $\forall_{i=j}: R_i = R_j$. since the top layer relevance values $R_j$ only need to be attributed towards one single respective input $i$ for each output neuron $j$.

After initiating the algorithm at the model output, it iterates over all the layers of the model towards the input, until relevance scores $R_i$ for all input components $x_i$ are obtained.
Assuming a (strong) positive model output represents the predicted presence of a class, then the input level relevance scores can be interpreted as follows:
Values $R_i \gg 0$ indicate components $x_i$  of the input which, due to the models’ learned decision function, represent the presence of the ``explained'' class, while conversely $R_i \ll 0$  contradict the prediction of that class.
$R_i \approx 0$  identify inputs $x_i$ which have no or only little influence to the model’s decision.

Applying above decomposition rules to a linear classifier $f(\x) = \sum_i x_iw_i + b$ with only a single output results in $z_{ij} = x_iw_i$ and $z_{bj} = b$, since the bias $b$ can be considered a constant always-on neuron, and $z_j = \sum_i x_iw_i + b = f(\x)$.
Initiating the (only) model output relevance as $R_j=f(\x)$ and substituting both $z_{ij}$ and $z_j$ in above relevance decomposition and pooling rules in Equations \eqref{eq:lrpdecomp} and \eqref{eq:lrppool} yields:
\begin{align}
R_{i\leftarrow j} & = \frac{x_iw_i}{f(\x)}f(\x) = x_iw_i 	&~;~\text{for inputs depending on data}~\x \nonumber\\
R_{b\leftarrow j} & = \frac{b}{f(\x)}f(\x) = b 				&~;~\text{for the relevance quantities for the bias}~b~.
\end{align}
Since the model only has one output, the pooling at each input $x_i$ (or the bias) becomes $R_i = R_{i \leftarrow j}$ (or $R_b = R_{b \leftarrow j}$).
In short, the application of LRP to a model consisting of only a single linear layer collapses to $R_i = x_iw_i$, the inherent explanation of the decision of a linear model in terms of input variables and the bias.
For further details, please refer to \cite{BachPLOS15}.

\subsection*{Data Availability}
The datasets generated and analysed during the current study are available in the Mendeley Data Repository~\cite{horst2018public} (\url{http://dx.doi.org/10.17632/svx74xcrjr.1}). The Layer-Wise Relevance Propagation Toolbox~\cite{LapJMLR16} (\url{https://github.com/sebastian-lapuschkin/lrp_toolbox}) and the experimental code derivation thereof is available on GitHub (\url{https://github.com/sebastian-lapuschkin/interpretable-deep-gait}).

\bibliographystyle{plain}
\bibliography{paper}

\section*{Acknowledgements}
The authors thank all the participating subjects for their time and patience as well as Christin Rupprecht and Eva Klein for her encouragement and support during the data collection.
No benefits in any form have been received or will be received from a commercial party related directly or indirectly to the subject of this article.
This work was supported by the Brain Korea 21 Plus Program through the National Research Foundation of Korea; the Institute for Information \& Communications Technology Promotion (IITP) grant funded by the Korea government (MSIT) [No. 2017-0-01779].
This work was also supported by the grant DFG (MU 987/17-1) and by the German Ministry for Education and Research as Berlin Big Data Centre (BBDC) (01IS14013A) and Berlin Center for Machine Learning under Grant 01IS18037I.
This publication only reflects the authors views.
Funding agencies are not liable for any use that may be made of the information contained herein.
This is a pre-print of an article
published in \emph{Scientific Reports}. The final authenticated version
is available online at: \url{https://doi.org/10.1038/s41598-019-38748-8}

\section*{Author Contributions}
FH and WIS conceived, designed and performed the experiment.
FH, SL, WS, KRM and WIS analysed the data.
SL, WS and KRM contributed analysis tools.
FH, SL, WS, KRM and WIS wrote the paper and drafted the article or revised it critically for important intellectual content.
All authors reviewed the manuscript.

\section*{Competing Interests}
The author(s) declare no competing interests.

\clearpage
\section*{SUPPLEMENTARY MATERIAL}

This document contains 3 supplementary tables with results in Tables \ref{tab:r1}-\ref{tab:r3}, 6 supplementary tables describing neural network architectures in Tables \ref{tab:a1}-\ref{tab:a6} and 4 supplementary subject specific analyses in Figures \ref{fig:sub21}-\ref{fig:sub56}.
\setcounter{figure}{0}
\setcounter{table}{0}
\renewcommand{\thetable}{ST\arabic{table}}
\renewcommand{\thefigure}{SF\arabic{figure}}
\renewcommand{\figurename}{Supplemental Material, Figure}
\renewcommand{\tablename}{Supplemental Material, Table}

\section*{Supplementary Results}

\begin{table}[h]
    \begin{center}
    \begin{tabular}{r|rrrrr}
    \textbf{Model}	&		\textbf{Ground} 			&		\textbf{Joint Angles}&		\textbf{Joint Angles}	&		\textbf{Joint Angles }	&		\textbf{Joint Angles}\\
            &		\textbf{Reaction}	&		\textbf{Full-Body}		&		\textbf{Full-Body}	 	&		\textbf{Lower-Body	}	&		\textbf{Lower-Body}\\
            &		\textbf{ Forces [\%]}			&		\textbf{[\%]}		&		\textbf{(flex.-ext.) [\%]}	&		\textbf{[\%]	}		&		\textbf{(flex.-ext.) [\%]}\\
    \hline
    Linear (SGD)&		95.4 (1.7)&		100.0 (0.0)&		96.3 (1.8)&		100.0 (0.0)&		91.5 (2.2)\\
    Linear (SVM)&		100.0 (0.0)&		100.0 (0.0)&		99.7 (0.4)&		100.0 (0.0)&		99.8 (0.6)\\
    MLP (2, 64)&		80.7 (3.6)&		100.0 (0.0)&		75.7 (4.7)&		99.9 (0.3)&		55.6 (6.5)\\
    MLP (2, 128)&		89.1 (2.9)&		100.0 (0.0)&		85.7 (4.7)&		99.9 (0.3)&		69.9 (5.7)\\
    MLP (2, 256)&		94.4 (2.3)&		100.0 (0.0)&		92.1 (3.2)&		100.0 (0.0)&		78.6 (4.2)\\
    MLP (2, 512)&		96.6 (0.8)&		100.0 (0.0)&		95.7 (1.2)&		100.0 (0.0)&		87.1 (2.6)\\
    MLP (2, 1024)&		98.6 (1.1)&		100.0 (0.0)&		97.0 (2.0)&		100.0 (0.0)&		91.9 (3.4)\\
    MLP (3, 64)&		88.3 (3.7)&		99.9 (0.3)&		84.3 (3.3)&		99.9 (0.3)&		68.5 (8.3)\\
    MLP (3, 128)&		92.5 (1.9)&		100.0 (0.0)&		90.9 (2.0)&		100.0 (0.0)&		84.2 (4.7)\\
    MLP (3, 256)&		96.6 (0.8)&		100.0 (0.0)&		95.6 (2.6)&		100.0 (0.0)&		89.4 (4.3)\\
    MLP (3, 512)&		98.2 (0.9)&		100.0 (0.0)&		97.0 (1.0)&		100.0 (0.0)&		94.0 (2.4)\\
    MLP (3, 1024)&		98.8 (1.3)&		100.0 (0.0)&		97.8 (1.0)&		100.0 (0.0)&		96.5 (1.2)\\
    CNN--A&		99.1 (0.8)&		100.0 (0.0)&		95.6 (1.7)&		99.9 (0.3)&		92.0 (3.9)\\
    CNN--A3&		99.6 (0.6)&		99.8 (0.4)&		95.0 (1.7)&		99.9 (0.3)&		94.2 (1.6)\\
    CNN--A6&		99.0 (0.8)&		99.9 (0.3)&		96.5 (1.3)&		99.8 (0.4)&		97.7 (1.4)\\
    CNN--C3&		99.2 (0.6)&		100.0 (0.0)&		97.7 (1.5)&		100.0 (0.0)&		97.0 (1.3)\\
    CNN--C3--3&		--&		100.0 (0.0)&		--&		99.9 (0.3)&		--\\
    CNN--C6&		99.2 (0.6)&		99.9 (0.3)&		97.2 (1.6)&		100.0 (0.0)&		95.6 (1.8)
    \end{tabular}
    \end{center}
    \caption{The classification rates of the subject-classification of different models using artificial neural networks (n=57), reported in pairs of $mean~(standard~deviation)$ in percent.}
    \label{tab:r1}
\end{table}

\begin{table}[h]
    \begin{center}
    \begin{tabular}{r|rrrrrrr}
    \textbf{Model / Noise}	&		\textbf{Gaussian}	&	\textbf{	Gaussian}&		\textbf{Gaussian }&		\textbf{Salt$^-$} 	&		\textbf{Pepper }	&\textbf{Salt$^+$ }	&		\textbf{Shot}\\
    &		 \textbf{Random} 	&		 \textbf{Random}&		 \textbf{Random}&		  \textbf{Random}	&		 \textbf{Random}	& \textbf{Random}	&		\textbf{Random}\\
        &		\textbf{$\sigma=0.5$}	&		\textbf{$\sigma=1$}	&		\textbf{$\sigma=2$}	&			&			&	&		\\
    \hline
    Linear (SGD)	&		40.5 (6.0)	&		45.6 (3.0)	&		47.6 (1.8)	&		46.1 (2.7)	&		37.1 (7.0)	&		46.4 (3.1)	&		46.5 (2.9)\\
    Linear (SVM)	&		7.4 (4.6)	&		26.2 (5.6)	&		39.9 (3.7)	&		34.0 (6.9)	&		5.0 (3.2)	&		35.9 (7.2)	&		35.9 (7.5)\\
    MLP (2, 64)	&		19.2 (15.9)	&		29.3 (13.6)	&		38.8 (9.3)	&		43.1 (7.3)	&		16.6 (15.8)	&		42.9 (7.6)	&		42.8 (7.7)\\
    MLP (2, 128)	&		14.4 (14.2)	&		25.9 (13.1)	&		36.8 (9.4)	&		42.0 (6.9)	&		11.7 (13.6)	&		41.8 (7.6)	&		41.7 (7.7)\\
    MLP (2, 256)	&		10.3 (12)	&		21.8 (12.4)	&		34.9 (9.6)	&		40.8 (6.7)	&		8.1 (11.1)	&		40.7 (7.2)	&		40.5 (7.2)\\
    MLP (2, 512)	&		7.6 (9.8)	&		19.0 (11.5)	&		32.5 (9.6)	&		39.6 (6.6)	&		5.7 (9.0)	&		39.4 (7.3)	&		39.5 (7.2)\\
    MLP (2, 1024)	&		5.9 (8.1)	&		17.0 (11.1)	&		31.9 (9.2)	&		38.8 (6.5)	&		4.4 (7.0)	&		38.4 (7.2)	&		38.6 (7.2)\\
    MLP (3, 64)	&		16.2 (14.0)	&		27.6 (12.6)	&		37.8 (9.0)	&		43.5 (7.1)	&		14.4 (13.6)	&		43.3 (7.4)	&		43.3 (7.4)\\
    MLP (3, 128)	&		12.5 (12.3)	&		24.3 (11.9)	&		35.8 (8.9)	&		42.7 (6.9)	&		10.3 (11.9)	&		42.4 (7.4)	&		42.5 (7.4)\\
    MLP (3, 256)	&		8.9 (9.7)	&		20.8 (10.9)	&		34.9 (8.4)	&		41.7 (6.8)	&		7.7 (9.5)	&		41.2 (7.4)	&		41.1 (7.5)\\
    MLP (3, 512)	&		6.9 (8.4)	&		18.7 (10.4)	&		33.4 (8.2)	&		40.8 (6.8)	&		5.5 (7.6)	&		40.4 (7.1)	&		40.4 (7.2)\\
    MLP (3, 1024)	&		5.4 (7.0)	&		16.7 (9.4)	&		32.4 (7.8)	&		40.0 (6.6)	&		4.3 (6.1)	&		39.4 (7.2)	&		39.4 (7.2)\\
    CNN--A	&		6.1 (6.8)	&		18.2 (9.4)	&		33.1 (8.0)	&		36.9 (6.6)	&		8.8 (8.5)	&		37.4 (7.5)	&		37.4 (7.6)\\
    CNN--A3	&		9.4 (7.7)	&		23.4 (9.0)	&		36.6 (7.4)	&		38.0 (7.4)	&		6.6 (6.8)	&		39.6 (6.7)	&		39.6 (6.7)\\
    CNN--A6	&		8.6 (7.4)	&		22.9 (9.3)	&		35.9 (7.6)	&		38.0 (7.0)	&		8.8 (8.3)	&		39.3 (7.0)	&		39.3 (6.9)\\
    CNN--C3	&		12.2 (8.6)	&		27.5 (9.4)	&		38.0 (7.0)	&		31.0 (8.5)	&		11.3 (8.6)	&		37.9 (8.2)	&		38.1 (8.1)\\
    CNN--C6	&		5.9 (6.5)	&		18.3 (9.3)	&		33.3 (7.8)	&		37.2 (6.6)	&		8.5 (8.2)	&		37.3 (7.2)	&		37.7 (7.2)

    \end{tabular}
    \end{center}
    \caption{The area over perturbation curve  of the subject-classification of ground reaction forces of different models using artificial neural networks for different noise perturbation runs (n=57).
    AOPC values are reported in pairs of $mean~(standard~deviation)$.
    Gaussian noise types add white noise additively to the input samples.
    Salt$^-$ and Salt$^+$ noise types set input variables to $-1$ and $1$ respectively.
    Pepper noise replaces input variables with $0$.
    Shot noise replaces input variables randomly chosen from the set $\lbrace -1,0,1 \rbrace$.
    }
    \label{tab:r2}
\end{table}

\begin{table}[h]
    \begin{center}
    \begin{tabular}{r|rrrrr}
    \textbf{Model}	&		\textbf{Ground} 			&		\textbf{Joint Angles}&		\textbf{Joint Angles}	&		\textbf{Joint Angles }	&		\textbf{Joint Angles}\\
            &		\textbf{Reaction}	&		\textbf{Full-Body}		&		\textbf{Full-Body}	 	&		\textbf{Lower-Body	}	&		\textbf{Lower-Body}\\
            &		\textbf{Forces}	&		\textbf{}		&		\textbf{(flex.-ext.)}	 	&		\textbf{}	&		\textbf{(flex.-ext.)}\\
    \hline
    Linear (SGD)	&		4.31 (0.25)	&		4.27 (0.08)	&		3.93 (0.12)	&		4.16 (0.10)	&		3.86 (0.15)\\
    Linear (SVM)	&		0.31 (0.08)	&		0.56 (0.10)	&		0.31 (0.07)	&		0.48 (0.09)	&		0.26 (0.05)\\
    MLP (2, 64)	&		1.61 (0.56)	&		2.47 (0.12)	&		1.96 (0.29)	&		2.11 (0.13)	&		1.89 (0.38)\\
    MLP (2, 128)	&		1.38 (0.46)	&		2.27 (0.10)	&		1.59 (0.25)	&		1.91 (0.11)	&		1.48 (0.32)\\
    MLP (2, 256)	&		1.12 (0.34)	&		2.09 (0.09)	&		1.36 (0.21)	&		1.74 (0.11)	&		1.21 (0.27)\\
    MLP (2, 512)	&		0.95 (0.28)	&		1.93 (0.08)	&		1.10 (0.15)	&		1.60 (0.10)	&		0.96 (0.21)\\
    MLP (2, 1024)	&		0.83 (0.21)	&		1.80 (0.08)	&		0.94 (0.12)	&		1.48 (0.09)	&		0.83 (0.18)\\
    MLP (3, 64)	&		1.31 (0.34)	&		2.73 (0.15)	&		1.58 (0.17)	&		2.37 (0.16)	&		1.41 (0.23)\\
    MLP (3, 128)	&		1.00 (0.23)	&		2.44 (0.14)	&		1.28 (0.13)	&		2.05 (0.14)	&		1.02 (0.17)\\
    MLP (3, 256)	&		0.84 (0.18)	&		2.21 (0.13)	&		1.05 (0.10)	&		1.86 (0.13)	&		0.85 (0.12)\\
    MLP (3, 512)	&		0.72 (0.15)	&		2.02 (0.11)	&		0.90 (0.08)	&		1.66 (0.11)	&		0.70 (0.10)\\
    MLP (3, 1024)	&		0.63 (0.12)	&		1.85 (0.10)	&		0.77 (0.08)	&		1.50 (0.10)	&		0.61 (0.09)\\
    CNN--A	&		0.30 (0.08)	&		0.56 (0.09)	&		0.35 (0.06)	&		0.49 (0.08)	&		0.32 (0.05)\\
    CNN--A3	&		0.37 (0.10)	&		0.64 (0.11)	&		0.37 (0.08)	&		0.55 (0.09)	&		0.35 (0.06)\\
    CNN--A6	&		0.32 (0.09)	&		0.64 (0.12)	&		0.38 (0.07)	&		0.58 (0.11)	&		0.27 (0.05)\\
    CNN--C3	&		0.35 (0.08)	&		0.50 (0.08)	&		0.43 (0.08)	&		0.48 (0.08)	&		0.44 (0.08)\\
    CNN--C3--3	&		--	&		2.74 (0.10)	&		--	&		2.32 (0.13)	&		--\\
    CNN--C6	&		0.60 (0.12)	&		0.59 (0.08)	&		0.45 (0.08)	&		0.62 (0.11)	&		0.27 (0.05)
    \end{tabular}
    \end{center}
    \caption{The coefficient of variation of the subject-classification of different models using artificial neural networks (n=57).
    Values are reported in pairs of $mean~(standard~deviation)$.}
    \label{tab:r3}
\end{table}

\clearpage
\section*{Neural Network Architectures}
Complete description of all evaluated artificial neural network architectures.
The complete set of features and their specific input dimensionalities $D = C \times T$ ($T$ = time points, $C$ = number of channels) for which results are reported in this study are:
\begin{itemize}
\item Ground Reaction Forces (GRF): $D = 6 \times 101$ 
\item Full-Body Joint Angles  (FBJA): $D = 33 \times 101$
\item Full-Body Joint Angles (flexion-extension) (FBJAX): $D = 10 \times 101$
\item Lower-Body Joint Angles (LBJA): $D = 18 \times 101$
\item Lower-Body Joint Angles (flexion-extension) (LBJAX): $D = 6 \times 101$
\end{itemize}
All multi-layer perceptron models (MLP ($\cdots$)) and Linear models are -- except for the number of input and hidden neurons -- uniform in structure for all evaluated feature sets.
Those models also receive the input as flattened arrays (vectorization by row concatenation) shaped $T \cdot C$. with $H$ hidden units per layer, of which the number is uniform across all layers of the model.
The variable $L$ describes the number of output labels.
The MLP and Linear architectures are summarized in Table \ref{tab:a1}.

\begin{table}[h]
    \begin{center}
    \small
    \begin{tabular}{r|rrrrrr}
    \textbf{Model}	&		\textbf{Layer 1	}&		\textbf{Layer 2}	&		\textbf{Layer 3}	&	\textbf{	Layer 4}	&		\textbf{Layer 5}	&	\textbf{	Layer 6}\\
    \hline
    Linear (SGD)	&		Dense (D,L)	&		 	&		 	&		 	&		 	&		 \\
    \hline
    MLP (2, H)	&		Dense (D,H)	&		ReLU	&		Dense (H,L)	&		SoftMax	&		 	&		 \\
    \hline
    MLP (3, H)	&		Dense (D,H)	&		ReLU	&		Dense (H,H)	&		ReLU	&		Dense (H,L)	&		SoftMax
    \end{tabular}
    \end{center}
    \caption{Model architectures for the linear (Linear (SGD)) and fully-connected (MLP ($\cdots$)) artificial neural networks.)}
    \label{tab:a1}
\end{table}
Where Dense ($N$,$M$) describes a linear / fully-connected / dense layer with $N$ input neurons and $M$ output neurons.

Due to construction, the output shape of a convolutional layer is determined via an interplay of layer parameters (filter size and stride parameters) and the input shapes.
Since above features are characterized in a varying number of data channels, the convolutional neural network (CNN) architectures have been adapted accordingly.
The following tables (Table \ref{tab:a2} - Table \ref{tab:a6}) will describe the architectures of the evaluated models in details, where convolutional layers will be described as
Conv($Fc$, $Ft$|$Sc$, $St|H$), where $Fc$ and $Ft$ are the channel axis span and a time axis span of the learned convolutional filter banks respectively, $Sc$ and $St$ are the channel axis stride and time axis stride and $H$ the number of learned filters ( = output depth).

The CNN models can be categorized in the following groups:
\begin{itemize}
\item CNN-A: Models which use one layer of convolutions and have square input filter sizes configured to read all channels at once and use the same scope to read from the time axis, \ie $Fc$ = $Ft$ = $C$.
\item CNN-A3: These models read all input channels at once ($Fc = C$), but the scope for reading the time axis is limited to $Ft = 6$. Two layers of convolutions.
\item CNN-A6: Same as CNN-A3, with the difference of $Ft = 3$ for the input layer.
\item CNN-C3: Uses convolution filters shaped ($Fc$, $Ft$) = ($3$,$3$)
\item CNN-C3-3: Uses convolution filters shaped ($Fc$, $Ft$) = ($3$,$3$) and a stride pattern of ($Sc$, $St$) = ($3$,$3$), \ie there is no overlap between locations where convolutional filters are applied. This model receives inputs padded with one additional columns of zeros, to extend the time axis to 102 entries.
\item CNN-C6: Uses convolution filters shaped ($Fc$, $Ft$) = ($6$,$6$)
\end{itemize}

\begin{table}[h]
    \begin{center}
    \small
    \begin{tabular}{r|rrrrrrrrr}
    \textbf{Model }	&		\textbf{Layer 1}	&		\textbf{Layer 2}	&		\textbf{Layer 3	}&		\textbf{Layer 4}	&		\textbf{Layer 5}	&		\textbf{Layer 6} 	&		\textbf{Layer 7}	&		\textbf{Layer 8}	&		\textbf{Layer 9}\\
    \hline
    CNN--A	&		Conv 	&		ReLU	&		Flatten	&		Dense	&		SoftMax	&		 	&		 	&		 & \\
    &		(6,6|1,1|32)	&			&			&		(3072,L)	&			&		 	&		 	&		 & \\
    &		&			&			&			&			&		 	&		 	&		 & \\
    \hline
    CNN--A3	&		Conv 	&		ReLU	&		Conv &		ReLU	&		Flatten	&		Dense 	&		SoftMax	& &		 \\
        &		 (6,3|1,1|32)	&			&		 (1,3|1,1|32)	&			&			&		 (3104,L)	&			& &		 \\
        &		 	&			&			&			&			&			&			& &		 \\
        \hline
    CNN--A6	&		Conv 	&		ReLU	&		Conv 	&		ReLU	&		Flatten	&		Dense	&		SoftMax	& &		 \\
    &		 (6,6|1,1|32)	&			&		 (1,6|1,1|32)	&			&			&		 (2912,L)	&			& &		 \\
    &			&			&			&			&			&		&			& &		 \\
    \hline
    CNN--C3	&		Conv	&		ReLU	&		Conv 	&		ReLU	&		Conv 	&		ReLU	&		Flatten	&		Dense 	&		SoftMax\\
    &		 (3,3|1,1|32)	&			&		 (3,3|1,1|32)	&			&		 (2,2|1,1|32)	&			&			&		 (3072,L)	&	\\
    &			&			&			&			&		&			&			&			&	\\
    \hline
    CNN--C3--3	&		–	&		–	&		–	&		–	&		–	&		–	&		–	&		–	&		–\\
    &			&			&			&			&		&			&			&			&	\\
    &			&			&			&			&		&			&			&			&	\\
    \hline
    CNN--C6	&		Conv 	&		ReLU	&		Flatten	&		Dense 	&		SoftMax	&		 	&		 	&		 & \\
    &		 (6,6|1,1|32)	&			&			&		 (3072,L)	&			&		 	&		 	&		 & \\
    &		 	&			&			&			&			&		 	&		 	&		 &
    \end{tabular}
    \end{center}
    \caption{Model architectures for the convolutional neural network architectures for ground reaction force variables (GRF).}
    \label{tab:a2}
\end{table}

\begin{table}[h]
    \begin{center}
    \small
    \begin{tabular}{r|rrrrrrrrr}
    \textbf{Model }	&		\textbf{Layer 1}	&		\textbf{Layer 2}	&		\textbf{Layer 3	}&		\textbf{Layer 4}	&		\textbf{Layer 5}	&		\textbf{Layer 6} 	&		\textbf{Layer 7}	&		\textbf{Layer 8}	&		\textbf{Layer 9}\\
    \hline
    CNN--A	&		Conv 	&		ReLU	&		Flatten	&		Dense&		SoftMax	&		 	&		 	&		 & \\
        &		 (33,33|1,1|64)	&			&			&		 (4416,L)	&			&		 	&		 	&		 & \\
            &	&			&			&		 	&			&		 	&		 	&		 & \\
    \hline
    CNN--A3	&		Conv	&		ReLU	&		Conv 	&		ReLU	&		Flatten	&		Dense 	&		SoftMax	& &		 \\
    &		 (33,3|1,1|64)	&			&		 (1,3|1,1|32)	&			&			&		 (3104,L)	&			& &		 \\
    &		&			&		&			&			&	&			& &		 \\
    \hline
    CNN--A6	&		Conv &		ReLU	&		Conv 	&		ReLU	&		Flatten	&		Dense 	&		SoftMax	& &		 \\
        &		 (33,6|1,1|32)	&			&		 (1,6|1,1|32)	&			&			&		 (2912,L)	&			& &		 \\
            &		 	&			&		 	&			&			&			&			& &		 \\
    \hline
    CNN--C3	&		Conv 	&		ReLU	&		Conv 	&		ReLU	&		Conv &		ReLU	&		Flatten	&		Dense	&		SoftMax\\
    &		 (3,3|1,1|32)	&			&		 (3,3|1,1|32)	&			&		 (3,3|1,1|16)	&			&			&		 (41040,L)	&		\\
            &		 	&			&		 	&			&			&			&			& &		 \\
    \hline
    CNN--C3--3	&		Conv 	&		ReLU	&		Conv	&		ReLU	&		Conv	&		ReLU	&		Flatten	&		Dense 	&		SoftMax\\
        &		 (3,3|3,3|64)	&			&		 (3,3|1,1|64)	&			&		 (3,3|1,1|32)	&			&			&		 (6720,L)	&		\\
            &		 	&			&		 	&			&			&			&			& &		 \\
    \hline
    CNN--C6	&		Conv 	&		ReLU	&		Conv 	&		ReLU	&		Conv	&		ReLU	&		Flatten	&		Dense 	&		SoftMax\\
        &		 (6,6|1,1|32)	&			&		 (6,6|1,1|32)	&			&		 (6,6|1,1|16)	&			&			&		 (24768,L)	&		\\
            &		 	&			&		 	&			&			&			&			& &
    \end{tabular}
    \end{center}
    \caption{Model architectures for the convolutional neural network architectures for full-body joint angles (FBJA).}
    \label{tab:a3}
\end{table}

\begin{table}[h]
    \begin{center}
    \small
    \begin{tabular}{r|rrrrrrrrr}
    \textbf{Model }	&		\textbf{Layer 1}	&		\textbf{Layer 2}	&		\textbf{Layer 3	}&		\textbf{Layer 4}	&		\textbf{Layer 5}	&		\textbf{Layer 6} 	&		\textbf{Layer 7}	&		\textbf{Layer 8}	&		\textbf{Layer 9}\\
    \hline
    CNN--A	&		Conv&		ReLU	&		Flatten	&		Dense	&		SoftMax	&		 	&		 	&		 & \\
        &		 (10,10|1,1|32)	&			&			&		 (2944,L)	&			&		 	&		 	&		 & \\
            &		 	&			&		 	&			&			&			&			& &		 \\
    \hline
    CNN--A3	&		Conv 	&		ReLU	&		Conv 	&		ReLU	&		Flatten	&		Dense 	&		SoftMax	& &		 \\
        &		 (10,3|1,1|32)	&			&		 (1,3|1,1|32)	&			&			&		 (3104,L)	&			& &		 \\
            &		 	&			&		 	&			&			&			&			& &		 \\
    \hline
    CNN--A6	&		Conv 	&		ReLU	&		Conv	&		ReLU	&		Flatten	&		Dense 	&		SoftMax	& &		 \\
    &		 (10,6|1,1|32)	&			&		 (1,6|1,1|32)	&			&			&		 (2912,L)	&			& &		 \\
            &		 	&			&		 	&			&			&			&			& &		 \\
    \hline
    CNN--C3	&		Conv 	&		ReLU	&		Conv&		ReLU	&		Conv 	&		ReLU	&		Flatten	&		Dense&		SoftMax\\
        &		 (3,3|1,1|32)	&			&		 (3,3|1,1|32)	&			&		 (3,3|1,1|32)	&			&			&		 (12160,L)	&		\\
            &		 	&			&		 	&			&			&			&			& &		 \\
    \hline
    CNN--C3--3	&		–	&		–	&		–	&		–	&		–	&		–	&		–	&		–	&		–\\
            &		 	&			&		 	&			&			&			&			& &		 \\
            &		 	&			&		 	&			&			&			&			& &		 \\
    \hline
    CNN--C6	&		Conv 	&		ReLU	&		Conv &		ReLU	&		Conv	&		ReLU	&		Flatten	&		Dense 	&		SoftMax\\
        &		 (6,6|1,1|32)	&			&		 (3,3|1,1|32)	&			&		 (3,3|1,1|32)	&			&			&		Dense (2944,L)	&		\\
            &		 	&			&		 	&			&			&			&			& &		 \\
    \end{tabular}
    \end{center}
    \caption{Model architectures for the convolutional neural network architectures for full-body joint angles in the sagittal plane (flexion-extension) (FBJAX).}
    \label{tab:a4}
\end{table}

\begin{table}[h]
    \begin{center}
    \small
    \begin{tabular}{r|rrrrrrrrr}
    \textbf{Model }	&		\textbf{Layer 1}	&		\textbf{Layer 2}	&		\textbf{Layer 3	}&		\textbf{Layer 4}	&		\textbf{Layer 5}	&		\textbf{Layer 6} 	&		\textbf{Layer 7}	&		\textbf{Layer 8}	&		\textbf{Layer 9}\\
    \hline
    CNN--A	&		Conv	&		ReLU	&		Flatten	&		Dense 	&		SoftMax	&		 	&		 	&		 & \\
        &		 (18,18|1,1|64)	&			&			&		 (5376,L)	&			&		 	&		 	&		 & \\
            &		 	&			&		 	&			&			&			&			& &		 \\
    \hline
    CNN--A3	&		Conv 	&		ReLU	&		Conv 	&		ReLU	&		Flatten	&		Dense 	&		SoftMax	& &		 \\
        &		 (18,3|1,1|64)	&			&		 (1,3|1,1|32)	&			&			&		 (3104,L)	&			& &		 \\
            &		 	&			&		 	&			&			&			&			& &		 \\
    \hline
    CNN--A6	&		Conv &		ReLU	&		Conv &		ReLU	&		Flatten	&		Dense 	&		SoftMax	& &		 \\
    &		 (18,6|1,1|64)	&			&		 (1,6|1,1|32)	&			&			&		 (2912,L)	&			& &		 \\
            &		 	&			&		 	&			&			&			&			& &		 \\
    \hline
    CNN--C3	&		Conv 	&		ReLU	&		Conv&		ReLU	&		Conv 	&		ReLU	&		Flatten	&		Dense 	&		SoftMax\\
    &		 (3,3|1,1|32)	&			&		 (3,3|1,1|32)	&			&		 (3,3|1,1|16)	&			&			&		 (18240,L)	&		\\
            &		 	&			&		 	&			&			&			&			& &		 \\
    \hline
    CNN--C3--3	&		Conv 	&		ReLU	&		Conv 	&		ReLU	&		Conv 	&		ReLU	&		Flatten	&		Dense	&		SoftMax\\
    &		 (3,3|3,3|64)	&			&		 (3,3|1,1|64)	&			&		 (3,3|1,1|32)	&			&			&		 (1920,L)	&		\\
            &		 	&			&		 	&			&			&			&			& &		 \\
    \hline
    CNN--C6	&		Conv 	&		ReLU	&		Conv &		ReLU	&		Conv 	&		ReLU	&		Flatten	&		Dense	&		SoftMax\\
        &		 (6,6|1,1|32)	&			&		 (6,6|1,1|32)	&			&		 (6,6|1,1|16)	&			&			&		 (4128,L)	&		\\
            &		 	&			&		 	&			&			&			&			& &		 \\
    \end{tabular}
    \end{center}
    \caption{Model architectures for the convolutional neural network architectures for lower-body joint angles (LBJA).}
    \label{tab:a5}
\end{table}

\begin{table}[h]
    \begin{center}
    \small
    \begin{tabular}{r|rrrrrrrrr}
    \textbf{Model }	&		\textbf{Layer 1}	&		\textbf{Layer 2}	&		\textbf{Layer 3	}&		\textbf{Layer 4}	&		\textbf{Layer 5}	&		\textbf{Layer 6} 	&		\textbf{Layer 7}	&		\textbf{Layer 8}	&		\textbf{Layer 9}\\
    \hline
    CNN--A	&		Conv 	&		ReLU	&		Flatten	&		Dense	&		SoftMax	&		 	&		 	&		 & \\
        &		 (6,6|1,1|32)	&			&			&		 (3072,L)	&			&		 	&		 	&		 & \\
            &		 	&			&		 	&			&			&			&			& &		 \\
    \hline
    CNN--A3	&		Conv 	&		ReLU	&		Conv 	&		ReLU	&		Flatten	&		Dense 	&		SoftMax	& &		 \\
        &		 (6,3|1,1|32)	&			&		 (1,3|1,1|32)	&			&			&		 (3104,L)	&			& &		 \\
            &		 	&			&		 	&			&			&			&			& &		 \\
    \hline
    CNN--A6	&		Conv 	&		ReLU	&		Conv	&		ReLU	&		Flatten	&		Dense 	&		SoftMax	& &		 \\
        &		 (6,6|1,1|32)	&			&		 (1,6|1,1|32)	&			&			&		 (2912,L)	&			& &		 \\
            &		 	&			&		 	&			&			&			&			& &		 \\
    \hline
    CNN--C3	&		Conv 	&		ReLU	&		Conv 	&		ReLU	&		Conv 	&		ReLU	&		Flatten	&		Dense &		SoftMax\\
        &		 (3,3|1,1|32)	&			&		 (3,3|1,1|32)	&			&		 (2,2|1,1|32)	&			&			&		 (3072,L)	&		\\
            &		 	&			&		 	&			&			&			&			& &		 \\
    \hline
    CNN--C3--3	&		–	&		–	&		–	&		–	&		–	&		–	&		–	&		–	&		–\\
            &		 	&			&		 	&			&			&			&			& &		 \\
            &		 	&			&		 	&			&			&			&			& &		 \\
    \hline
    CNN--C6	&		Conv 	&		ReLU	&		Flatten	&		Dense 	&		SoftMax	&		 	&		 	&		 &  \\
        &		 (6,6|1,1|32)	&			&			&		 (3072,L)	&			&		 	&		 	&		 &  \\
            &		 	&			&		 	&			&			&			&			& &		 \\
    \end{tabular}
    \end{center}
    \caption{Model architectures for the convolutional neural network architectures for lower-body joint angles in the sagittal plane (flexion-extension) (LBJAX).}
    \label{tab:a6}
\end{table}

\clearpage
\section*{Supplementary Subject Specific Analyses}

\begin{figure}[h]
\begin{center}
\includegraphics[width=.495\textwidth]{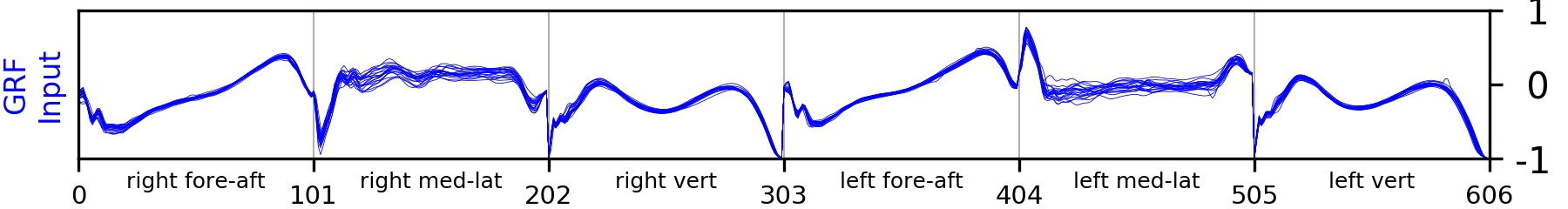}
\includegraphics[width=.495\textwidth]{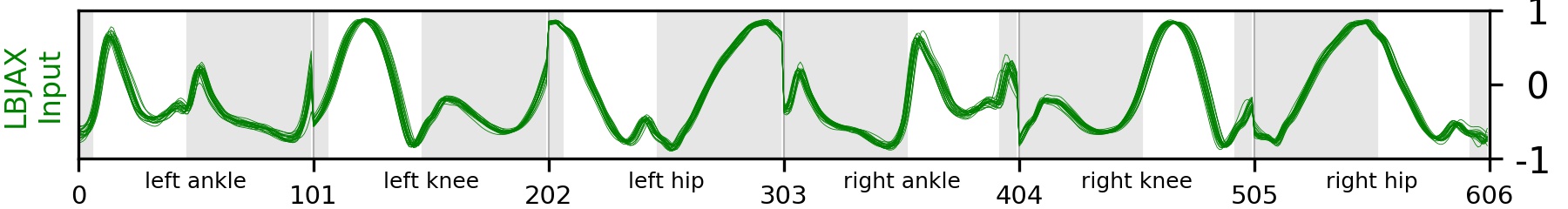}\\
\includegraphics[width=.495\textwidth]{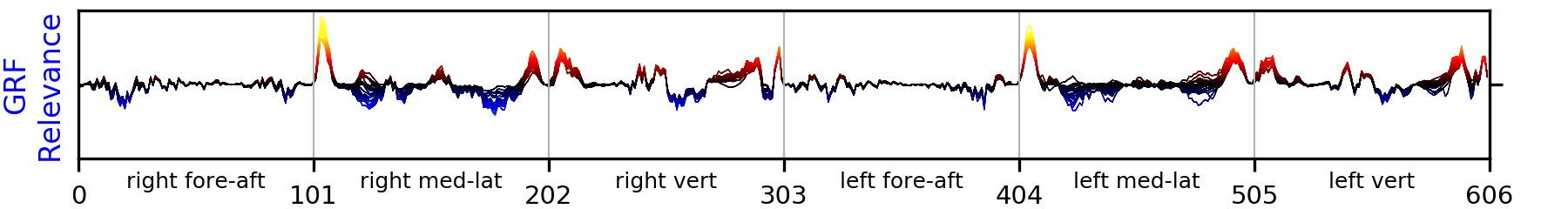}
\includegraphics[width=.495\textwidth]{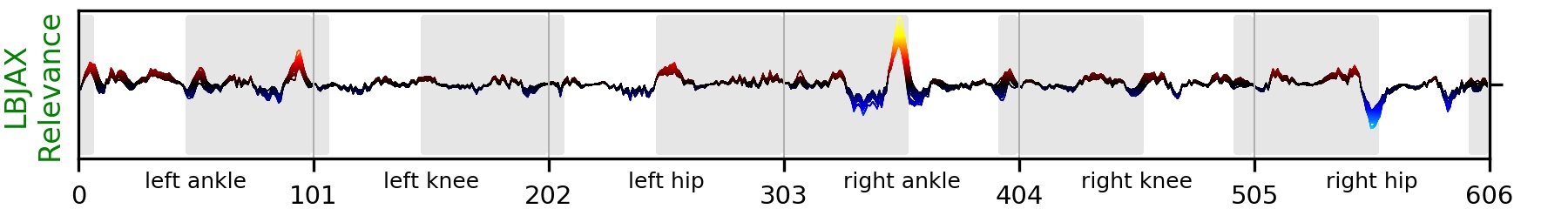}
\end{center}
\caption{Subject 21: GRF and JLBAX inputs as line plots \emph{(top)}.
The corresponding relevance maps \emph{(bottom)} for the CNN-A, following the procedure outlined in Figure 1 in the paper.
\emph{Left:} Ground reaction force (GRF).
The input relevance values point out that the model has learned to identify subject 21 by the medial-lateral shear force during the initial contact of the stance phase.
The GRF input balues reveal that the medial-lateral shear forces during the initial contact of the stance phase are among the highest in the sample for both foot contacts of subject 21.
\emph{Right:} Lower-body joint angles in the sagittal plane (flexion-extension) (LBJAX).
It can be observed from the input relevance values that the extension of the ankle is unique to subject 21 during the terminal stance phase of the right foot, which is identified by the model as an identifying characteristic.
For this subject, the model achieves TP rates of $100\%$ for LBJAX and $95.23\%$ for GRF.
}
\label{fig:sub21}
\end{figure}

\begin{figure}[h]
    \begin{center}
    \includegraphics[width=.495\textwidth]{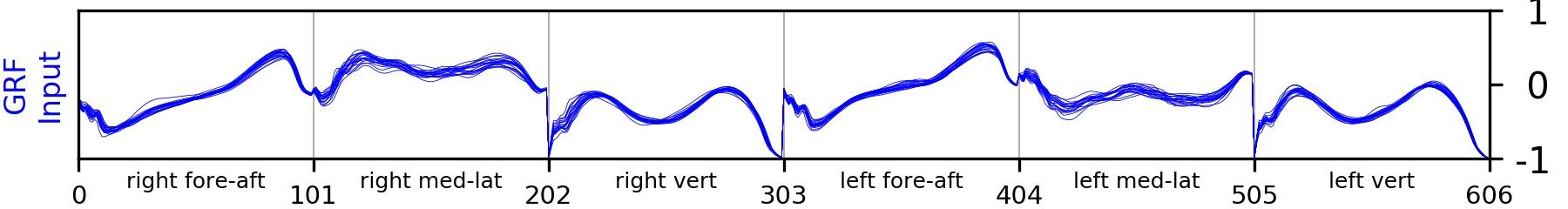}
    \includegraphics[width=.495\textwidth]{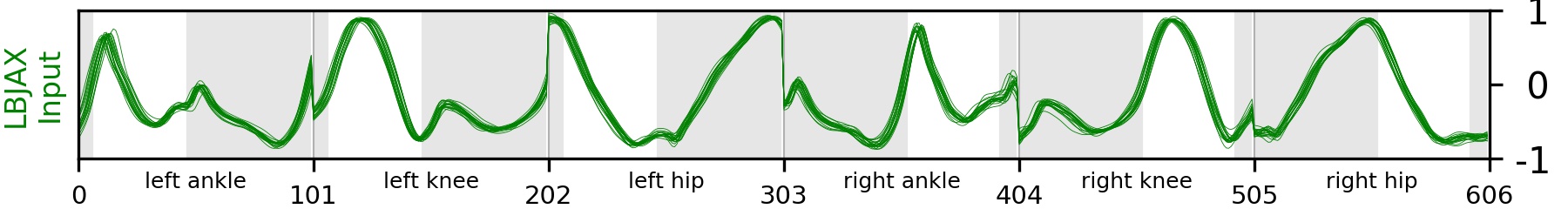}\\
    \includegraphics[width=.495\textwidth]{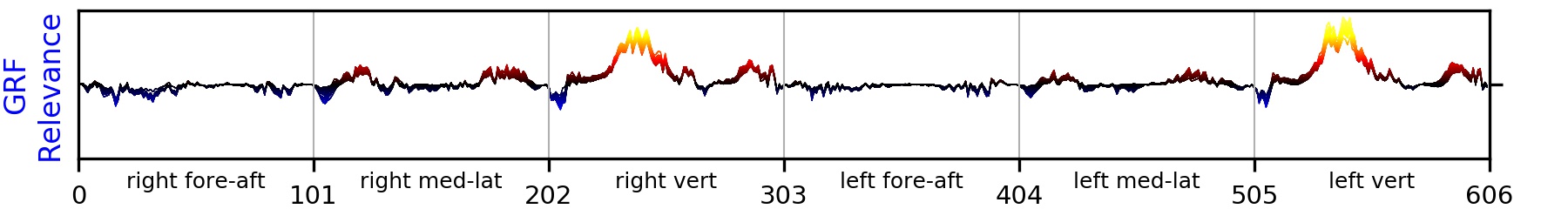}
    \includegraphics[width=.495\textwidth]{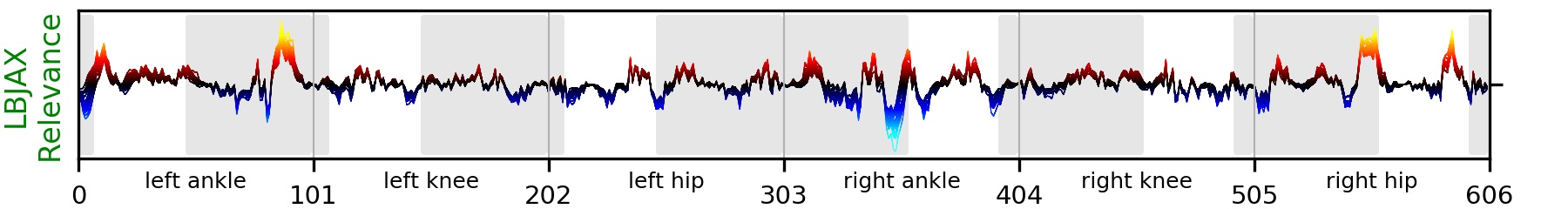}
    \end{center}
    \caption{Subject 50: GRF and JLBAX inputs as line plots \emph{(top)}.
    The corresponding relevance maps \emph{(bottom)} for the CNN-A, following the procedure outlined in Figure 1 in the paper.
    \emph{Left:} Ground reaction force (GRF).
    The highest input relevance values are appearning for the vertical GRF during the mid stance phase of the right and left foot contact.
    This indicates that the time window of features, where the leg is located directly over the foot and the other leg is in the middle of the swing phase, is unique to subject 50.
    \emph{Right:} Lower-body joint angles in the sagittal plane (flexion-extension) (LBJAX).. From the input relevance values, we can observe that the extension of the ankle during the terminal stance phase of the left leg as well as the flexion of the hip joint during the terminal stance phase and the mid swing phase of the right leg is unique to subject 50.
    For this subject, the model achieves TP rates of $100\%$ for GRF but only $80\%$ for LBJAX, for which it consistently mispredicts as subject 4.
    The model's uncertainty is reflected in the comparatively noisy and piece-wise negative relevance values.
    Relevance feedback such as for subject 50 and LBJAX data, providing detailed insight into the reasoning behind the model's uncertain prediction,
     yields information critical for the practical applicability of gait analysis tools based on machine learning in a clinical setting.
    }
    \label{fig:sub50}
\end{figure}

\begin{figure}[h]
    \begin{center}
    \includegraphics[width=.495\textwidth]{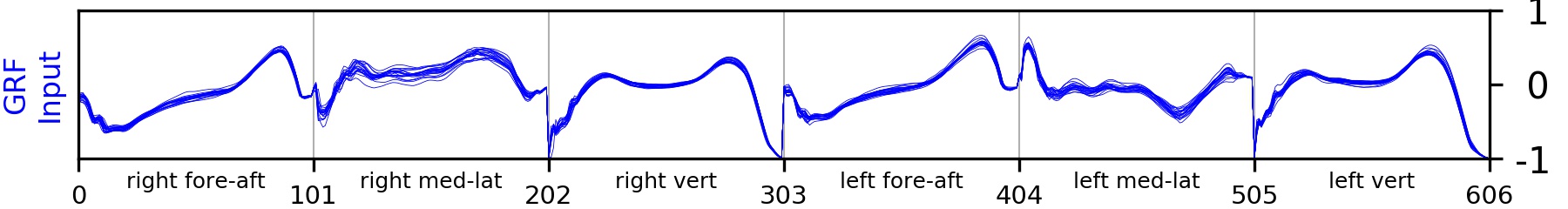}
    \includegraphics[width=.495\textwidth]{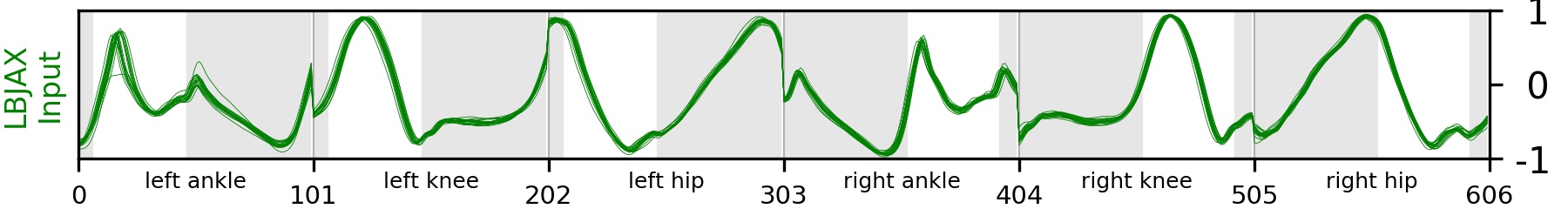}\\
    \includegraphics[width=.495\textwidth]{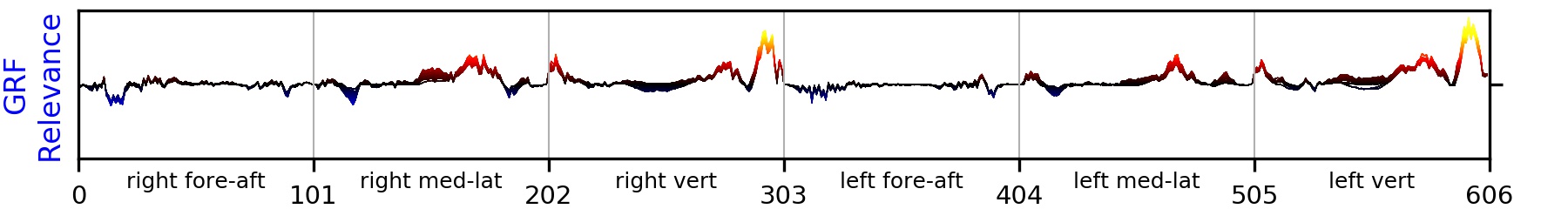}
    \includegraphics[width=.495\textwidth]{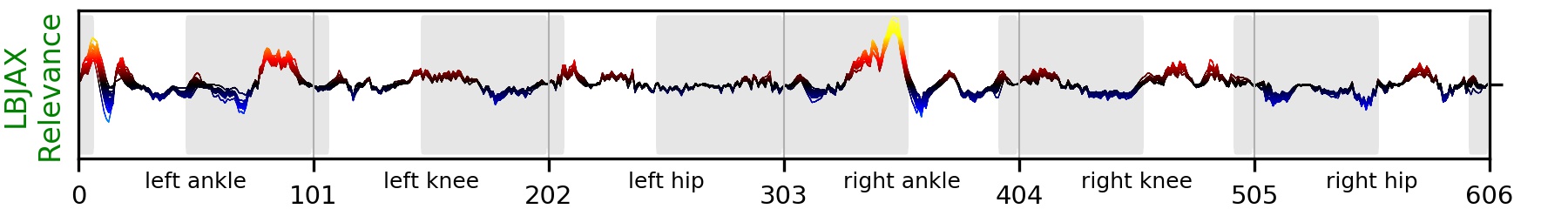}
    \end{center}
    \caption{Subject 54: GRF and JLBAX inputs as line plots \emph{(top)}.
    The corresponding relevance maps \emph{(bottom)} for the CNN-A, following the procedure outlined in Figure 1 in the paper.
    \emph{Left:} Ground reaction force (GRF).
     The input relevance values point out that the model has learned to identify subject 54 based the vertical GRF during the terminal stance phase of the right and left leg.
    \emph{Right:} Lower-body joint angles in the sagittal plane (flexion-extension) (LBJAX).
    The LBJAX supports the finding about the unique nature of the terminal stance phase to subject 54, showing the highest input relevance values for the extension of the ankle during this time window of the gait cycle.
    For this subject, the model achieves TP rates of $95\%$ for LBJA and $100\%$ for GRF.}
    \label{fig:sub54}
\end{figure}

\begin{figure}[h]
    \begin{center}
    \includegraphics[width=.495\textwidth]{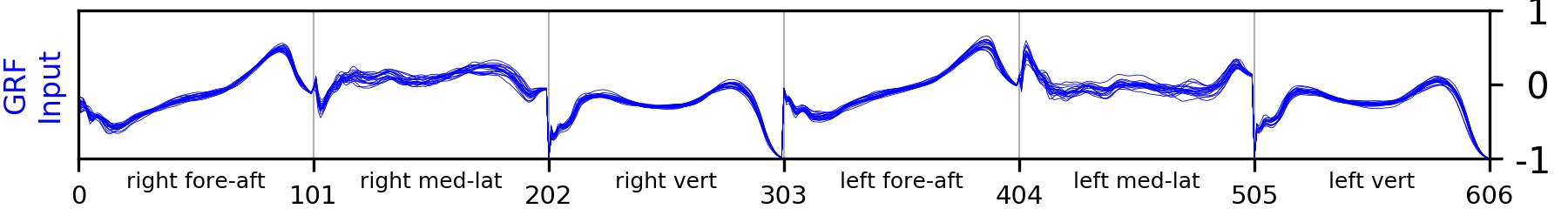}
    \includegraphics[width=.495\textwidth]{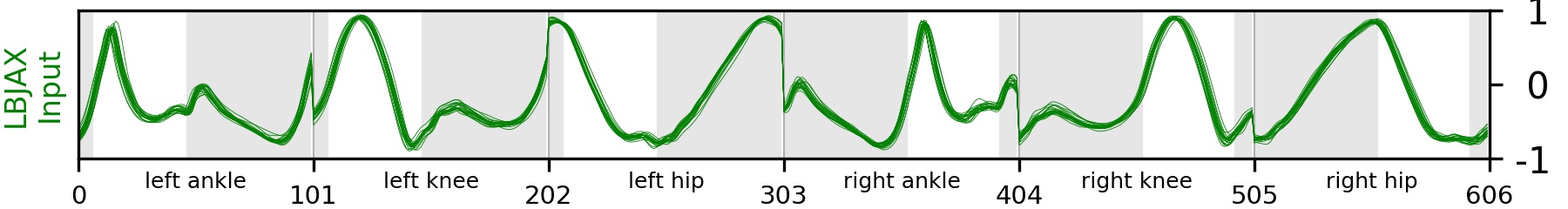}\\
    \includegraphics[width=.495\textwidth]{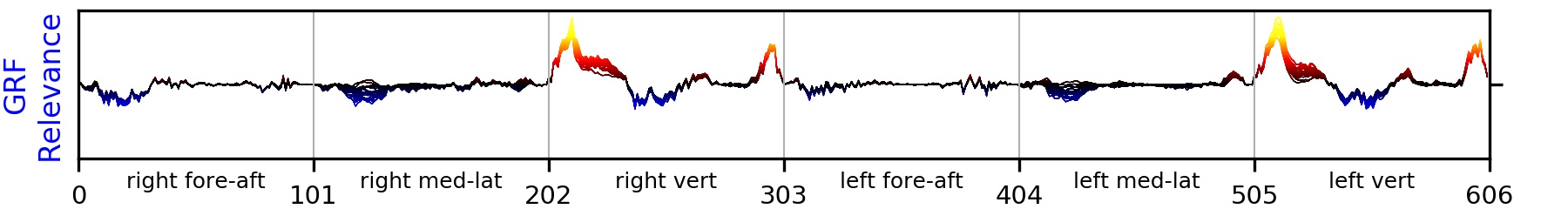}
    \includegraphics[width=.495\textwidth]{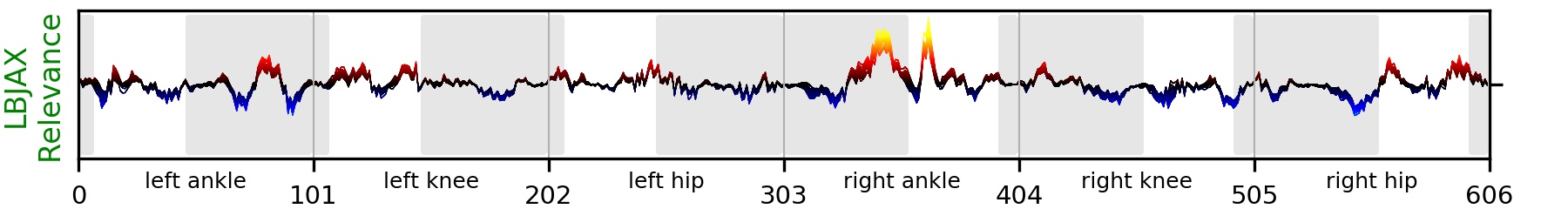}
    \end{center}
    \caption{Subject 56: GRF and JLBAX inputs as line plots \emph{(top)}.
    The corresponding relevance maps \emph{(bottom)} for the CNN-A, following the procedure outlined in Figure 1 in the paper.
     \emph{Left:} Ground reaction force (GRF).
    From the input relevance values, we can observe that the vertical GRF during the initial and terminal
    stance phase of the right and left leg is unique to subject 56.
     \emph{Right:} Lower-body joint angles in the sagittal plane (flexion-extension) (LBJAX).
     From the input relevance values, we can observe that the extension of the ankle during the terminal stance phase of the right is unique to subject 56.
    For this subject, the model achieves TP rates of $100\%$ for both LBJAX and GRF.}
    \label{fig:sub56}
\end{figure}

\end{document}